\documentclass[sigconf]{acmart}
\usepackage{notoccite}
\usepackage{subcaption}
\usepackage{algorithm}
\usepackage{algorithmic}
\usepackage{appendix}
\usepackage{multirow}
\usepackage{breqn}
\usepackage{amsmath}
\usepackage{booktabs,caption}
\usepackage[flushleft]{threeparttable}
\AtBeginDocument{%
  }

\setcopyright{acmlicensed}
\copyrightyear{2025}
\acmYear{2025}
\acmDOI{XXXXXXX.XXXXXXX}

\acmJournal{JACM}
\acmVolume{37}
\acmNumber{4}
\acmArticle{111}
\acmMonth{8}

\begin{document}

\title{Pretraining Large Brain Language Model\\
for Active BCI: Silent Speech}

\author{Jinzhao Zhou, Zehong Cao, Yiqun Duan, Connor Barkley, Daniel Leong, Xiaowei Jiang, Quoc-Toan Nguyen, Ziyi Zhao, Thomas Do, Yu-Cheng Chang, Sheng-Fu Liang, Chin-Teng Lin}

\makeatletter
\let\@authorsaddresses\@empty
\makeatother
\begin{abstract} 
This paper explores silent speech decoding in active brain-computer interface (BCI) systems, which offer more natural and flexible communication than traditional BCI applications. We collected a new silent speech dataset of over 120 hours of electroencephalogram (EEG) recordings from 12 subjects, capturing 24 commonly used English words for language model pretraining and decoding. Following the recent success of pretraining large models with self-supervised paradigms to enhance EEG classification performance, we propose Large Brain Language Model (LBLM) pretrained to decode silent speech for active BCI. To pretrain LBLM, we propose Future Spectro-Temporal Prediction (FSTP) pretraining paradigm to learn effective representations from unlabeled EEG data. Unlike existing EEG pretraining methods that mainly follow a masked-reconstruction paradigm, our proposed FSTP method employs autoregressive modeling in temporal and frequency domains to capture both temporal and spectral dependencies from EEG signals. After pretraining, we finetune our LBLM on downstream tasks, including word-level and semantic-level classification. Extensive experiments demonstrate significant performance gains of the LBLM over fully-supervised and pretrained baseline models. For instance, in the difficult cross-session setting, our model achieves 47.0\% accuracy on semantic-level classification and 39.6\% in word-level classification, outperforming baseline methods by 5.4\% and 7.3\%, respectively. Our research advances silent speech decoding in active BCI systems, offering an innovative solution for EEG language model pretraining and a new dataset for fundamental research. 
\end{abstract}


\received{20 February 2007}
\received[revised]{12 March 2009}
\received[accepted]{5 June 2009}

\begin{teaserfigure}
    \centering
    \includegraphics[width=0.83\textwidth]{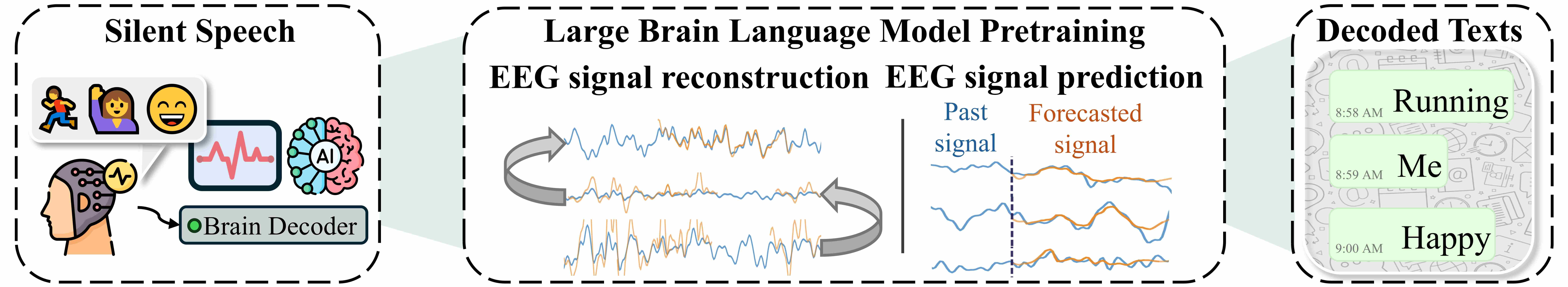}
    \caption{The proposed of silent speech active BCI paradigm. Users' neural activity captured during silent articulations is decoded into words to enable more natural communication. }\label{fig:cover}
    \vspace{0em}
  \end{teaserfigure}


\maketitle
\section{Introduction}  
Brain-computer interface (BCI) applications have been growing rapidly in recent years~\cite{pan2024comprehensive}. Among them, non-invasive electroencephalography (EEG) device a the competitive choice to build up BCI applications due to the low-risk protocol and affordability~\cite{wolpaw2013brain,saha2021progress,le2021theoretical}. Compared to passive BCIs like SSVEP~\cite{wang2016benchmark}, which require a user to focus on flickers to activate certain commands, active BCIs allow users to voluntarily control and interact with the surrounding environment, positioning as a potential alternative to manual or touchless input devices~\cite{lance2012brain,van2012brain}. For instance, motor imagery (MI)-based active BCI systems convert imagined limb movements into directional control signals~\cite{palumbo2021motor,pfurtscheller2001motor}. However, MI-based active BCI remains counterintuitive and limits the user's ability to express more complex ideas. In contrast, language is inherently more natural and versatile, allowing for more efficient and expressive communication. Relative to speech recognition that relies on audible sounds, silent speech in active BCI systems holds significant potential for assisting users with speech impairments or those who need to communicate in quiet environments.


EEG has been widely used for decoding brain activities in language processing~\cite{proix2022imagined,huth2012continuous,graham2013decoding}. To decode text from EEG signals, recent efforts such as EEG2text~\cite{wang2022open_aaai_eeg2text} and DeWave~\cite{duan2023dewave_brain2text} align EEG signals to text to generate sentences the participant reads. On the other hand, brainmagic~\cite{defossez2023decoding} decodes perceived speech from EEG by training a model to predict the speech spectrogram that the participant is listening to. Despite their success, two key limitations continue to hinder the advancement of language-based active BCIs:

\textbf{(1) The difference between passive and active EEG-based BCIs in language decoding:} 
Passive BCI paradigms for text generation rely on external stimuli, such as audio or visual contents, to create brain patterns corresponding to listening or reading, which may not be practical outside lab environments~\cite{de2024imagined,musch2020transformation,cheung2016auditory,broderick2018electrophysiological,brennan2019hierarchical,hollenstein2018zuco}. In contrast, silent speech active BCI decodes neural patterns that does not depend on external stimuli. Aside from application considerations, the difference in cognitive processing indicates that models trained on passive EEG data are not suitable to be used for active EEG decoding. Moreover, existing datasets for silent speech are typically small and often consist of non-meaningful utterances~\cite{dasalla2009single} or have a limited vocabulary set~\cite{nieto2022thinking,zhao2015classifying}, which further restricts their applicability in real-world scenarios and in training large brain models.

\textbf{(2) Lack of self-supervised training paradigms for EEG silent speech:} Over the years, we have witnessed the growth of self-supervised training methods across various domains which outcompete traditional superivsed learning methods. However, in EEG-based language decoding, self-supervised training methods remains underexplored. While multimodal EEG alignment methods have proven effective for bootstrapping EEG representation learning~\cite{D_fossez_2023_meg_eeg_clip_pretrain_meta_brain2speech,zhou2024belt}, they rely on pairing EEG signals with external modalities, such as visual or auditory stimuli presented simultaneously,which may not be feasible in silent speech. In contrast, unimodal self-supervised methods can leverage unlabeled EEG data to learn rich representations without external modalities, offering greater flexibility, particularly in language-related EEG tasks where labeled data is often scarce. 
However, most existing self-supervised methods for non-language EEG rely on reconstructing masked encoded features~\cite{jiang2024large,wang2024eegpt}. While effective to some extent, their methods do not require models to capture how EEG signals evolve over time, which may produce less meaningful and less effective representations for decoding the highly complex dynamics of silent speech EEG.

In this study, we address the issues discussed by developing the pretrained Large Brain Language Model (LBLM). First, we collected a large-scale silent speech dataset with over 120 hours of recordings from 12 subjects. Our dataset focuses on a vocabulary of 24 commonly used English words, such as `running', `happy', and `home', from six semantic groups. Then, we propose the Future Spectro-Temporal Prediction (FSTP), a self-superivsed pretraining paradigm to pretrain the LBLM model in an autoregressive fashion. The FSTP paradigm explicitly models both the temporal dependencies of EEG signals and their corresponding spectral components that are crucial for language decoding. Specifically, the FSTP paradigm consists of two stages where we first pretrain the LBLM backbone using the Masked Spectro-Temporal Prediction (MSTP) method, and then the Autoregressive Spectro-Temporal Prediction (ASTP) pretraining method is used to learn more complex temporal trends and frequency patterns by predicting future EEG waves and frequency components. We pretrain our LBLM model with over 22 million parameters on the collected dataset using a conformer backbone with layer-gating mechanism. For downstream classification tasks, a spatio-temporal classifier is developed to integrate and select backbone representations from different channels. The main contributions of this paper are:

\begin{itemize}  
  \item \textbf{Self-supervised pretraining paradigm for silent speech}: We propose a self-supervised pretraining method for EEG backbone models that does not require labeled data or external modalities. This pretraining paradigm encourages models to capture temporal-spectral patterns from EEG signals, enabling them to effectively perform downstream tasks. 
  \item \textbf{Large Brain Language Model (LBLM) for Silent Speech Decoding}: We propose LBLM, a 22-million-parameter model for feature extraction from language-related EEG signals. Extensive experiments evaluate the effectiveness of fine-tuning LBLM on word-level and semantic-group classification tasks. Our results show that LBLM achieves state-of-the-art performance in cross-session evaluations. 
  \item \textbf{Large-Scale Silent Speech EEG Dataset}: We collected a 120-hour EEG dataset from 12 participants for silent speech decoding, with 16 sessions recorded from each participant. The dataset will be made publicly available to support active BCI research and neural pattern analysis.  
\end{itemize}  

\section{Related Work}
\subsection{EEG Semantic Decoding}
EEG decoding has significantly enhanced our understanding of how the human brain processes semantic concepts~\cite{stansbury2013natural,graham2013decoding,proix2022imagined,huth2012continuous}. Earlier works on this topic mainly focused on decoding less meaningful syllables~\cite{zhao2015classifying,tottrup2019decoding,deng2010eeg} or directional words~\cite{van2021inner,koizumi2018development,cooney2020evaluation} via fully-supervised learning. For complex sentence decoding tasks, while sophisticated models were used, EEG encoders were still treained end-to-end from scratch without pretraining. For example, Défossez et al.~\cite{defossez2023decoding} trained a convolutional neural networks (CNNs)-based EEG encoder by aligning EEG spectrograms with those of audio signals. Similarly, EEG-to-Text methods~\cite{wang2022open_aaai_eeg2text,duan2023dewave_brain2text} paired a transformer-based EEG encoder with a pretrained language model (LM) to generate sentences. In their approaches, only the language model is pretrained, while the EEG is randomly initialized. Although these methods can generate a complete sentence thanks to the pretrained LM, we found that the EEG encoder is not trained properly to produce discriminative representations that are sufficient for accurately classifying words or sentences. Consequently, the generated sentences could not to reflect the meanings conveyed by the EEG inputs~\cite{jo2024eeg,zhou2024towards}. To this end, a training method tailored to learning effective EEG representations is needed to move EEG semantic decoding beyond its current bottleneck.

Recently, self-supervised pretraining methods for EEG have emerged to improve the generalizability of learned EEG representations. For instance, LaBraM~\cite{jiang2024large} and EEGPT~\cite{wang2024eegpt} have leveraged masked reconstruction pretraining on large-scale EEG datasets to learn general EEG representations. However, the neural tokenizer proposed in LaBraM mainly learns frequency domain features while the EEGPT focus on improving feature quality in the time domain. The uni-domain representation learned by their methods may be effective in silent speech decoding where patterns changes both temporally and spectrally. On the other hand, due to the potential information leakage from bidirectional attention in their masked-reconstruction pretraining methods, models may rely on surrounding EEG tokens to guess the masked ones without learning the underlying temporal or frequency dynamics of the EEG signals. Different from their methods, the proposed FSTP method moves beyond their limitation by pretraining an EEG model to predict future EEG wave signals and their frequency components in an autoregressive setting. By removing future information from the model input, the FSTP method forces the model to learn the temporal dependency and more meaningful representations. Moreover, by additionally predicting future frequency components, we further enhance LBLM by learning time-frequency features from the EEG signals for effective silent speech decoding. 

\subsection{Future Time-series Prediction}
Our pretraining method is inspired by recent advances in time-series prediction models, as EEG signals can be naturally viewed as multivariate, non-stationary time-series with complex temporal dynamics. For predicting complex future time-series, earlier methods such as Temporal convolutional neural (TCN)~\cite{hewage2020temporal}, N-BEATS~\cite{oreshkin2019n}, and deep autoregressive recurrent networks (DeepAR)~\cite{salinas2020deepar} exemplify using neural networks to capture short-term temporal patterns. However, these models could not capture longer-range dependencies. To do so, Temporal Fusion Transformer (TFT)~\cite{lim2021temporal} integrates attention mechanisms for multi-horizon forecasting, while PatchTST~\cite{nie2022time} uses patch-based tokenization for better long-term dependency modeling. More recently, utilizing massive amount of time-series data, TimesFM~\cite{das2023decoder} demonstrates autoregressive training on time-series data learns more generalized representation for predicting longer sequences in the future. Despite their success, these time-series prediction models are not immediately applicaable for EEG decoding. This is because these model are not immediately applicable for EEG decoding mainly because their architectures are not designed for classification tasks, and they are only trained to capture temporal patterns from non-EEG datasets without learning frequency-domain features that are critical to understanding EEG signals.

In our work, we frame future time-series prediction as a self-supervised learning method, where a model can learn more diverse features to the EEG signal without labels. Additionally, unlike existing time-series prediction methods that only model the time-series signal in the time domain, our FSTP method is tailored for EEG to predict future EEG signals as well as its future frequency power to enhance the diversity of the learned EEG representations for downstream classification tasks.


\section{Method} 
\begin{figure*}[h!]
    \centering
    \includegraphics[width=0.9\linewidth]{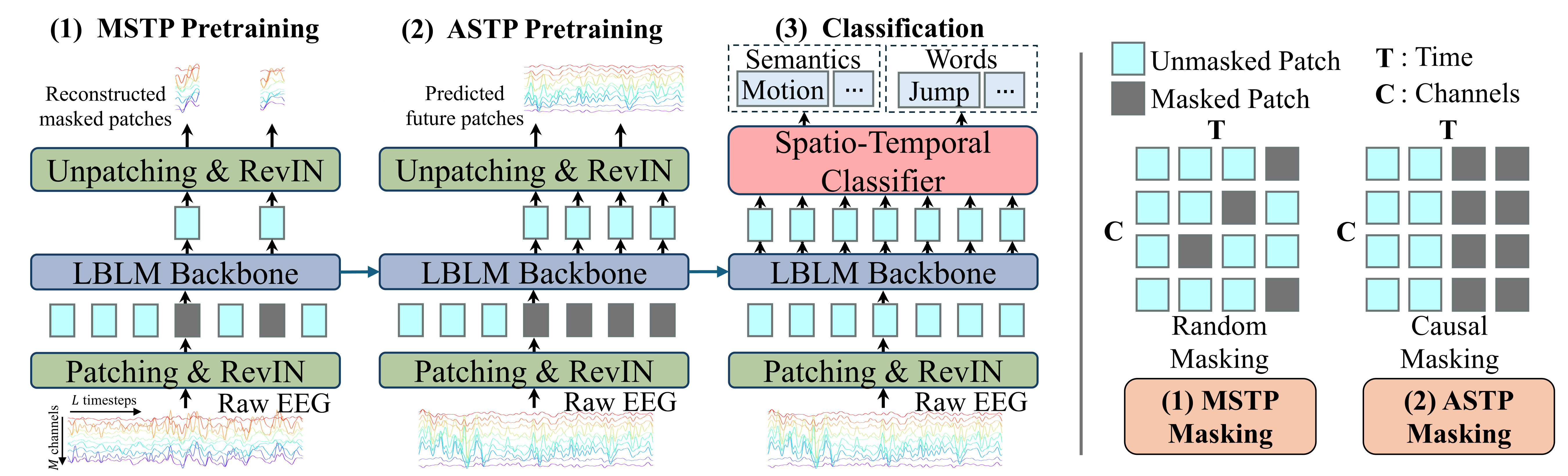}
    \caption{
    Overview of the proposed Future Spectro-Temporal Prediction (FSTP) pretraining paradigm for the LBLM backbone, which consists of two stages. In Stage (1), Masked Spectro-Temporal Prediction (MSTP) pretraining is used to warm up the model weights by reconstructing randomly masked EEG wave and spectra. In Stage (2), future EEG patches are completely masked out in Autoregressive Spectro-Temporal Prediction (ASTP) to help the model learn non-trivial representations. For classification, a spatio-temporal classifier is added to aggregate and select the representations learned by the LBLM backbone. We finetune the whole model for effective semantic-level and word-level classification tasks. \label{fig:main}
    }
    \vspace{-1em}
\end{figure*}
In this section, we present the proposed training paradigm and the detailed backbone architecture for LBLM. We consider the input EEG data to be a multivariate time-series with $L$ timesteps. $\mathbf{x}^(m) = \{x_1,\cdots, x_t, \cdots, x_L\}$ represents signals from the $m^{th}$ channels, $1\leq m\leq M$. To learn effective EEG representations for silent speech decoding, we construct our LBLM model using conformer blocks and pretrain by the proposed FSTP paradigm.

As shown in Figure \ref{fig:main}, we first pretrain the backbone of the LBLM using the MSTP method, where the input EEG signals are randomly masked out, and the backbone model learns contextualized representations to reconstruct the masked EEG waves as well as the spectrum. We then further train the backbone using the ASTP method to force the model to predict future EEG states in both temporal and spectral domains from more meaningful representations such as long-range temporal dynamics. Finally, we finetune the pretrained backbone with a spatio-temporal classification head for both semantic-level and word-level classification tasks. 

\subsection{EEG Input Patching} 
The aim of our LBLM model backbone is to understand the temporal dynamics of EEG data with a non-stationary nature. To avoid the information bottlenecks introduced by Vector Quantized (VQ) encoders~\cite{wang2024eegpt,jiang2024large}, we use EEG patches from each channel as input tokens to the LBLM backbone model. In particular, we divide EEG input $\mathbf{x}^(m)$ into a sequence of overlapping patches $\{\mathbf{p}_1,\cdots,\mathbf{p}_N\}$with the patch length of $P$ and an overlapping stride of $S$. Consequently, each input EEG channel will be segmented into $N=\lfloor\frac{L-P}{S}+1\rfloor$ patches. In our experiment, we downsampled the EEG signal to $250\mathrm{Hz}$ and set $P=25$ to enable the model to capture local patterns within $100ms$ while setting $S=6$ to maintain strong correlations between consecutive patches.

\subsection{LBLM Backbone Model}
\subsubsection{Position and Subject Embedding} 
\begin{figure}[h!]
    \centering
    \includegraphics[width=0.85\linewidth]{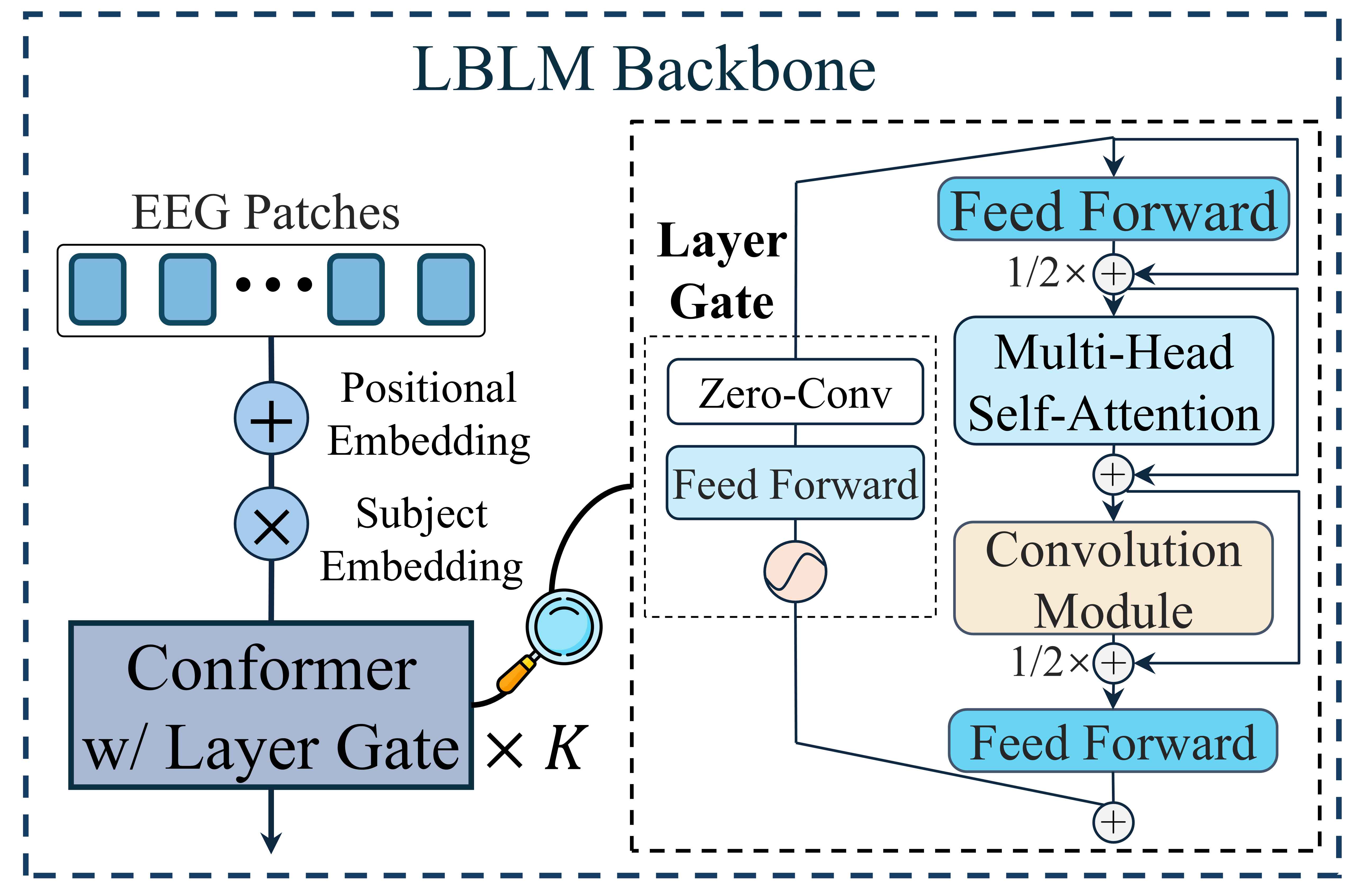}
    \caption{
        The proposed LBLM backbone. EEG signals are first segmented into overlapping patches and embedded with positional and subject embeddings. We build the backbone model using Conformer blocks with a layer-gating mechanism to stabilize training. The gating connects the input and output tokens using a zero convolution layer with $1\times{1}$ convolution with both weight and bias initialied to zero.
\label{fig:model-backbone}
    }     
    \vspace{-1em}
\end{figure}

As shown in Figure \ref{fig:model-backbone}, we enhance the input EEG tokens by incorporating positional and subject embeddings into the input patches. For positional encoding, we initialize the embedding list $PE=\{\mathrm{pe}_1, \cdots, \mathrm{pe}_N\}$ using sinusoidal encoding to inject the sequential information of the EEG patches \cite{vaswani2017attention}. To account for inter-subject variability and differences in electrode impedance, we further multiply the input token by a subject-specific embedding $SE=\{\mathrm{se}_1, \cdots, \mathrm{se}_K\}$ for all $K$ subjects in the dataset. The subject embedding vector is initialized with all-one values to allow the model to gradually learn subject-dependent transformations. Thus, we update a EEG token to $\tilde{\mathbf{p}}$ by: 
\begin{equation}
    \{\tilde{\mathbf{p}}=(\mathbf{p}_i+\mathrm{pe}_i)\times{\mathrm{se}_k}| i=1,\cdots,N, k=\mathrm{subj}(\mathbf{p})\}
\end{equation}
where $\mathrm{subj}(\mathbf{p})$ returns the subject number of the input patches.

\subsubsection{Backbone Architecture} 
Next, the embedded patch will be fed to our LBLM backbone model, which is built by Conformer blocks~\cite{gulati2020conformer}. Different from a Transformer block~\cite{vaswani2017attention}, Conformer adds a convolution module after the multi-head self-attention (MHSA) layer. We will present more details on this convolution module in Appendix \ref{appendix:ConvolutionModule}. Similar to the change of pitch and phonemes that can be captured within a small time window to convey important information in speech, short-term fluctuation in EEG, such as Event-Related Potentials (ERPs), Event-related desynchronization (ERD), and microstates, is also considered as basic units carrying information about certain cognitive processes. Empirically, we find the Conformer to be more effective for processing EEG because the kernel sliding and neighbor aggregation mechanism allow it to capture the mentioned basic units as well as the occurrence order. On the other hand, the MHSA layer focuses on capturing the global dependencies among all EEG patches by the following dot-product attention mechanism: 
\begin{equation}
    \mathrm{Softmax}(\frac{\tilde{\mathbf{p}}^\top{W^Q}(\tilde{\mathbf{p}}^\top{W^K})^\top}{\sqrt{d}})\tilde{\mathbf{p}}^\top{W^V}  \vspace{-1em}
\end{equation}
where $\sqrt{d}$ is a scaling factor. $W^K$, $W^V$, and $W^Q$ are linear projections for transforming $\tilde{\mathbf{p}}$ into query, key, and value representations. Finally, the layer input is transformed into a $d$-dimensional output token $\mathbf{e}$. 

To make the training of the Conformer model more stable, we incorporate a layer-gating mechanism to the Conformer block as shown in Figure \ref{fig:model-backbone}. Specifically, we connect the input token from the previous layer $\mathbf{e}^{l-1}$ with the output token $\mathbf{e}^{l}$ after the Conformer transformation. 
\begin{equation}
    \mathbf{e}^{l} = g(\mathbf{e}^{l-1})\mathbf{e}^{l} + (1-g(\mathbf{e}^{l-1}))\mathbf{e}^{l-1} 
\end{equation}
where $g(\mathbf{e}^{l-1})$ is a gating function implemented by a submodule comprising a zero convolution layer, a feed-forward layer, and a sigmoid function, generating token-wise gating values between 0 and 1. The zero convolution layer follows the design from~\cite{zhang2023adding}, where we use a $1\times{1}$ convolution with both weight and bias initialized to zero. This gating mechanism ensures the more progressive integration of information from previous layers and prevents performance degradation due to network depth. For downstream tasks such as classification, the gating mechanism also enables feature selection from earlier network layers to avoid over-reliance on the features from the final Conformer layer, which could be overfitted to a pretraining task. 

\subsection{Backbone Pretraining}\label{sec:FSTP}
As shown in Figure \ref{fig:main}, we propose to pretrain the LBLM backbone using the FSTP paradigm, which enhances the learned representations by predicting future EEG signals in both time and frequency domains. The FSTP paradigm consists of two stages: (1) the MSTP method, which predict the masked EEG patches and their spectral components simultaneously, and (2) the ASTP method, which autoregressively predicts future EEG patches and spectral components which encourage the model to learn more diverse and robust representations from unlabeled EEG data.

\subsubsection{Token-wise Prediction Head} 
Unlike most time-series signals, EEG signals are characterized by high stochasticity and nonstationarity, making it significantly more difficult to reconstruct the original signal and predict future signals. Although most time-series forecasting models \cite{zhou2021informer,nie2022time} flatterns all tokens into a massive $N\times{d}$ vector and apply a prediction head to the flatterned vector, in our previous experiments, we found that the loss fails to converge and the predicted future EEG signals failed to reflect any meaningful continuation of past patterns or trends. We consider it is because the flattening operation leads to a significant loss of the sequential structure of the EEG signals, and the large layer weights make the training less stable~\cite{wang2021understanding,roscacontinuous}. Therefore, we propose to use a token-wise prediction head for our FSTP pretraining as shown in Figure \ref{fig:multi-head}, following the design of decoder-only language models~\cite{radford2019language}, without flattening the output tokens. Except for improved training stability, our token-wise prediction design offers several advantages: (1) it reduces the number of parameters from $N \times d \times T$ to $d \times T$, (2) eliminates the need to reinitialize the prediction head when changing the target prediction length $T$, and (3) enables flexible switching between the MSTP and ASTP stages without requiring separate prediction heads for each pretraining method.

\subsubsection{Spectro-Temporal Prediction}
\begin{figure}[h!]
    \centering
    \begin{subfigure}[b]{0.44\linewidth}
        \centering
        \includegraphics[width=\linewidth]{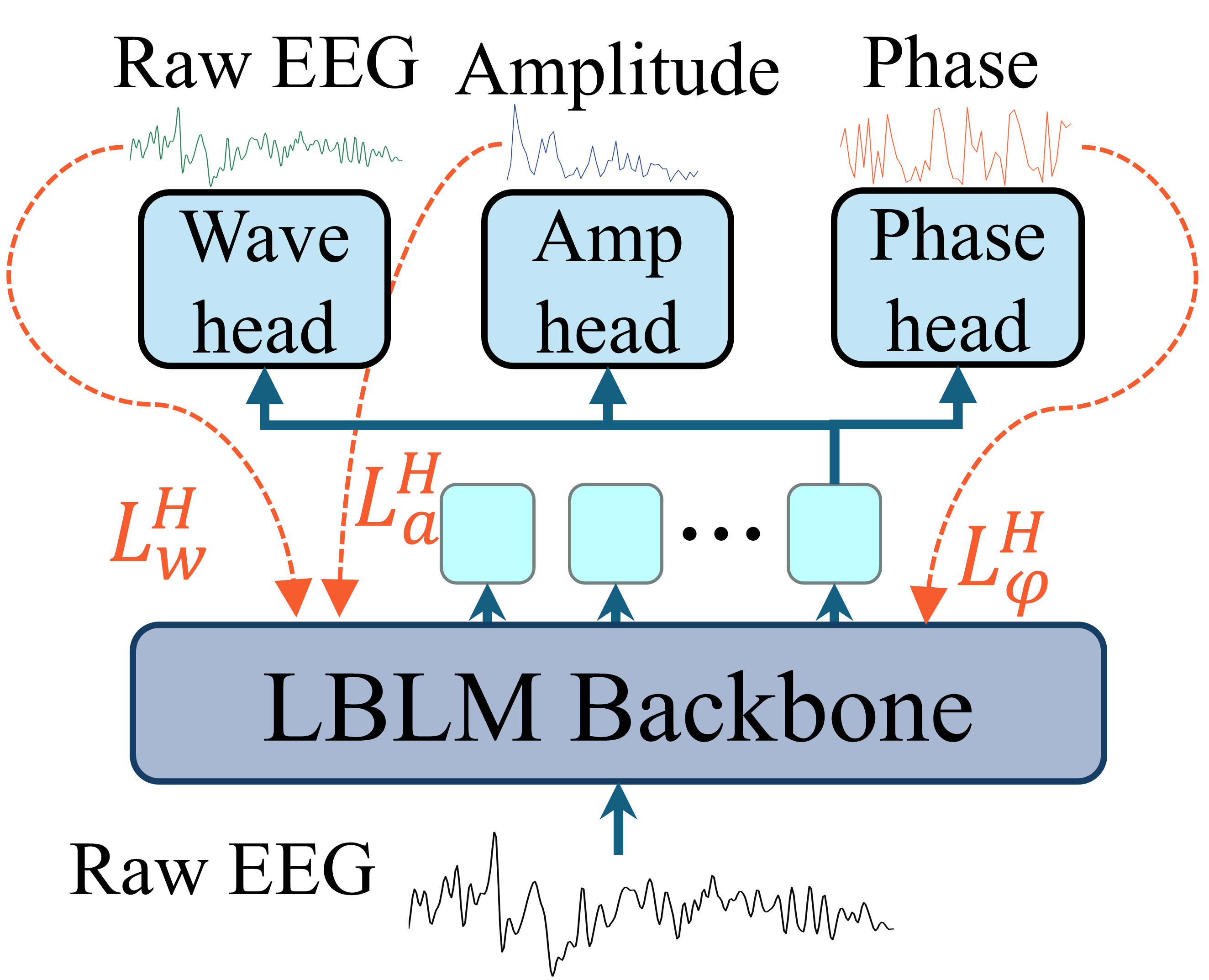}
        \caption{Token-wise Prediction Head in FSTP. \label{fig:multi-head}}
       
    \end{subfigure}
    \begin{subfigure}[b]{0.52\linewidth}
        \centering
        \includegraphics[width=\linewidth]{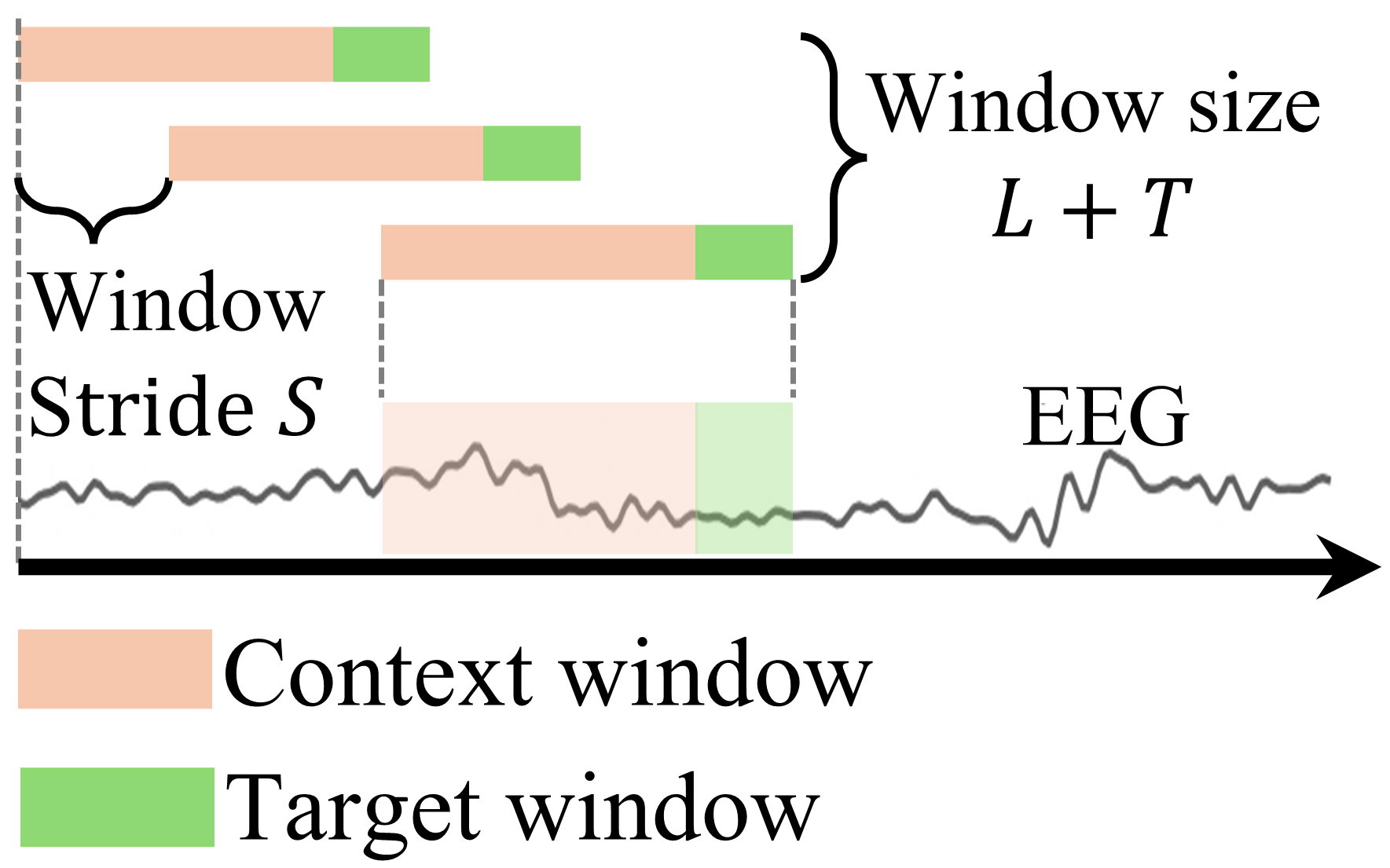}
        \caption{EEG sequence sampling during ASTP.\label{fig:sliding-window}
    }        
    \end{subfigure}
    \caption{%
    (a) In FSTP pretraining, the backbone model predicts the raw EEG wave, the Fourier amplitude and the Fourier phase from each output token of the LBLM backbone. A separate prediction head for each modality.  
    (b) Sampling the input context and target window during ASTP pretraining.
    }    
    \vspace{-1em}
\end{figure}

We propose pretraining a model to understand both the temporal and spectrum aspects of the EEG signals. As shown in Figure \ref{fig:multi-head}, we create $3$ token-wise prediction heads for the prediction of EEG wave, Fourier amplitude, and Fourier phase. For brevity, we present the calculation of these Fourier components in Appendix \ref{appendix:fft}. During the MSTP and FSTP pretraining, we use a Huber loss $L^H$ to reduce sensitivity to noise and sudden changes of amplitude in EEG signals during training~\cite{gokcesu2021generalized}. The Huber loss can be written as:
\begin{equation}
    L^{H}(\mathbf{x}, \hat{\mathbf{x}},\delta) = 
    \begin{cases} 
    \dfrac{1}{2} \, (\mathbf{x} -\hat{\mathbf{x}})^2 & \text{if } |\mathbf{x}_p -\hat{\mathbf{x} }| \le \delta, \\ 
    \delta \,\Bigl(|\mathbf{x} -\hat{\mathbf{x}}| - \tfrac{1}{2} \delta \Bigr) & \text{if } |\mathbf{x} -\hat{\mathbf{x}}| > \delta
    \end{cases}
    \label{eq:huber}
\end{equation}
where $\mathbf{x}_p$ and $\hat{\mathbf{x}_p}$ denote the ground truth and predicted values for the EEG wave, amplitude or phase. $\delta$ is a hyperparameter that controls the transition between quadratic and linear loss terms. 

\subsubsection{Masked Spectro-Temporal Prediction Pretraining}
To learn generic representations from the EEG data and warmup the model weights, we first perform MSTP pretraining as shown in Figure \ref{fig:main}-(1). Here, we randomly masked out $r$ percentage of the input token using a mask $\mathcal{M}$ and pretrain the LBLM backbone to reconstruct the masked EEG wave, amplitude and phase. Denotes $\mathbf{P}^u$ to be all unmasked EEG patches, $\mathbf{P}^m$ to be all masked patches. We calculate the Fourier amplitude and Phase of the masked as $\mathbf{A}^m$ and $\pmb{\phi}^m$. The training objective function can be written as: 
\begin{dmath}
    \mathcal{L}_{MSTP} = \frac{1}{|\mathcal{M}|}\sum^{|\mathcal{M}|}_{i=1}{L^H_w(f^{w}(\mathbf{P}^u),\mathbf{P}_i^m)+\lambda_1 L^H_a(f^{a}(\mathbf{P}^u),\mathbf{A}_i^m) + \lambda_2 L^H_\phi(f^{\phi}(\mathbf{P}^u),\pmb{\phi}_i^m)}
    \label{eq:loss-MSTP}
\end{dmath}
where $f^{w}$, $f^{a}$, and $f^{\phi}$ denotes the prediction head for the wave, amplitude, and the phase,e respectively. We add a subscript to the  Huber Loss terms to signify applying the Huber loss to a specific domain. For reconstructing the amplitude and phase, have weighting coefficients $\lambda_1$, $\lambda_2$ to $L^H_a$ and $L^H_\phi$ to control their impact on the overall training objective.

\subsubsection{Autoregressive Spectro-Temporal Prediction Pretraining}
We further propose the ASTP pretraining to enhance the features extracted by LBLM backbone. Recent self‐supervised EEG approaches~\cite{jiang2024large,wang2024eegpt} only train a backbone model to reconstruct the masked tokens. However, we consider that due to the strong temporal continuity of EEG signals, masking contiguous overlapping regions could lead to trivial prediction solutions such as interpolation, averaging, or using the value from the most adjacent patches. Different from their pretraining method, our ASTP method enforce strict autoregressive setting to the pretraining and is the first method to show the feasibility of training a EEG backbone model to predict future EEG trends. As shown in Figure \ref{fig:sliding-window}, ASTP training requires the backbone model to predict the raw EEG waves and their spectra components in the $T$ future time steps given a context window of $L$ time steps. Reusing the Huber loss term, the objective function for ASTP can be written as: 
\begin{dmath}
    \mathcal{L}_{ASTP} =L^{H}_w(\mathbf{P}_{<L},\mathbf{P}_{L:L+T}) +\lambda_1 L^{H}_a(\mathbf{P}_{<L},\mathbf{A}_{L:L+T})+ \lambda_2 L^{H}_\phi(\mathbf{P}_{<L},\pmb{\phi}_{L:L+T})  
    \label{eq:loss-FSTP}
\end{dmath}
where Huber loss is applied to compute the difference between the actual and predicted waveforms ($L^{H}_w$), amplitude $L^{H}_a$, and phase $L^{H}_\phi$. Since amplitude and phase jointly characterize the underlying EEG signal, optimizing them together leads to representations that better reflect the underlying temporal structure and spectral properties of the EEG data. 

\begin{figure}[h!]
    \centering
    \includegraphics[width=0.8\linewidth]{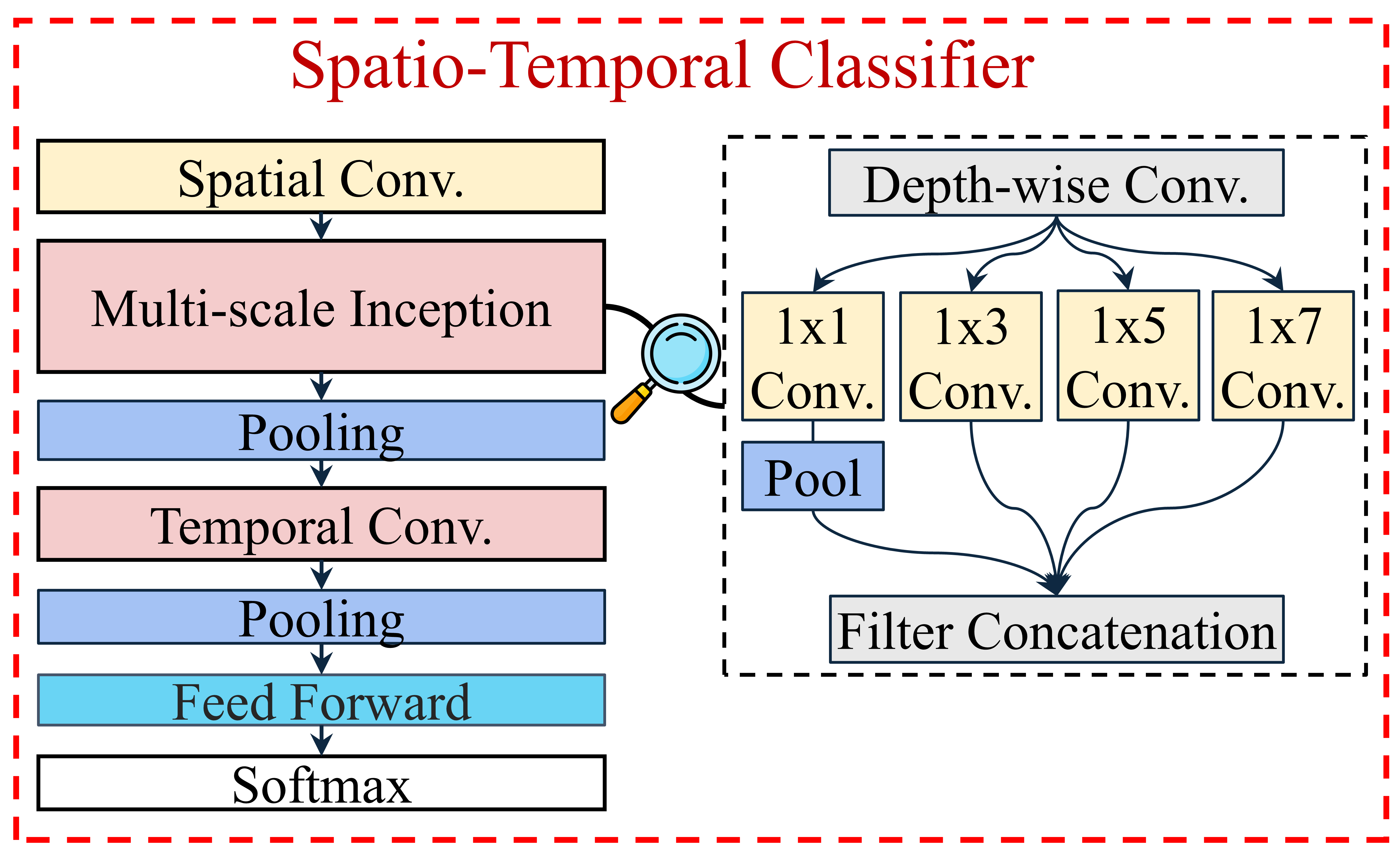}
    \caption{Detailed structure of the Spatio-Temporal (ST) classifier for downstream tasks exploiting EEG patterns learned from backbone LBLM. The ST classifier integrates features across EEG channels and extracts multi-scale temporal patterns to improve classification performance.\label{fig:model-cls}}
    \vspace{-1em}
\end{figure}
\subsection{Spatio-Temporal Classifier}\label{sec:st-cls} 
For the downstream classification task, we use a classifier that leverages the feature learned by the LBLM backbone. Drawing from prior work on EEG decoding~\cite{li2021automatic,ingolfsson2020eeg,hersche2018fast,wang2008local,lawhern2018eegnet}, we incorporate multi-scale temporal filtering and spatial aggregation as essential components for capturing spatio-temporal patterns. As illustrated in Figure~\ref{fig:model-cls}, our Spatio-Temporal Classifier begins with a spatial convolution that aggregates information across all EEG channels to extract topographical patterns relevant to silent speech. Then, two temporal modules, a multi-scale inception block and a temporal convolution block, are used to capture both short- and long-range temporal dependencies. The multi-scale inception block enhances subject adaptability by processing short-term features at multiple receptive field sizes simultaneously, using convolutional kernels of size $1\times 1$, $1\times 3$, $1\times 5$, and $1\times 7$. The subsequent temporal convolution block integrates longer-term temporal context by further combining and refining these features. Finally, all representations are pooled and passed through a feedforward layer with a softmax activation to generate classification probabilities.

\section{Experiments}
\begin{figure}[h!]
    \centering
    \includegraphics[width=0.8\linewidth]{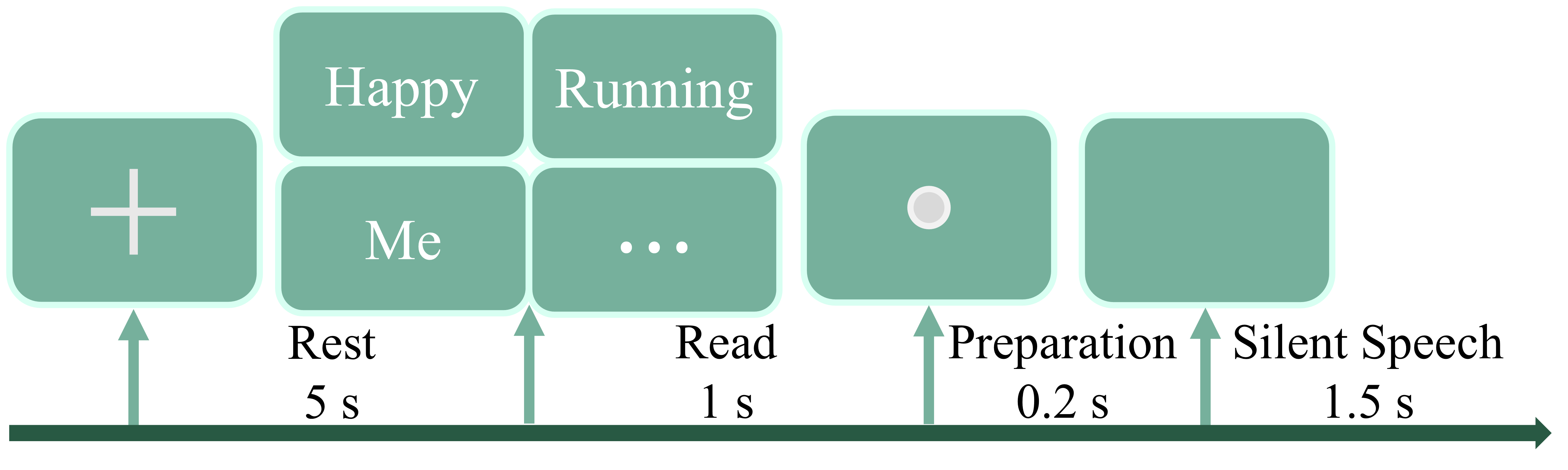}
    \caption{
        EEG-based Silent Speech Experiment design. Each trial consisted of four segments: rest, read, preparation, and silent speech. The number in between indicates the duration of each segment. \label{fig:experiment-setup}}
        \vspace{-1em}
\end{figure}
\subsection{Data Collection}
We recruited $12$ healthy individuals (S1-S10) for our EEG-based silent speech experiment. The experimental protocol is shown in Figure \ref{fig:experiment-setup} and consists of four segments: rest, read, preparation, and silent speech production. The rest segment lasts for $5$ seconds, showing an eye fixation cross (+) $1.5$ seconds before the word cue appear. In the read segment, the participants were presented with a cue and were asked to read the word presented on the screen. The read segment lasts for $1$ second. At the end of the read segment, an audio cue signals the transition to the silent speech segment for a brief preparation period of $0.2$ second. Afterwards, participants were given $1.5$ seconds to silently produce the same word. To reduce fatigue, participants are provided with a breaks after every $20$ trials for up to 5 minutes. Additional details regarding data collection procedures, equipment, preprocessing and analysis can be found in Appendix~\ref{sec:dataset-collection}. Altogether, we have collected $6000$ trials from each participant with $16$ EEG sessions. Each session contains $375$ silent speech trials. The duration for an entire session last for approximately $40$ minutes including the break period. The duration of the whole dataset is approximately $120$ hours in total. 

\subsection{Training and Evaluation}
\subsubsection{Pretraining and Finetuning}
We divided the data from each subject into training, validation, and testing splits in a cross-session way, where 2 sessions from each participant were held out from the 16 sessions during the pretraining and finetuning stages. For the classification task, one held-out session was used for validation and model selection, while the other was reserved exclusively for testing, and all reported results are based on the test session. In total, we utilized approximately 105 hours of unlabeled EEG recordings for the pretraining stage and over 5,200 labeled silent speech segments for finetuning. 

During pretraining, we train our LBLM backbone by MSTP and FSTP tasks outlined in Section \ref{sec:FSTP}. Reversible normalization (RevIN)~\cite{kim2021reversible} is applied to normalize the input data and reconstruct predictions in the original scale, while multi-band dataset mixing is used to enlarge the training set and enhance diversity during training. Please refer to Appendix \ref{appendix:RevIN} and \ref{appendix:Mixing} for more details on RevIN and dataset mixing used in our paper. For finetuning on semantic- and word-level classification, we employ cross-entropy loss, which is described in Appendix \ref{appendix:CE}. Implementation details, as well as hyperparameter settings, are described in Appendix \ref{appendix:hyperparameter}.  

\subsubsection{Baselines and Metrics}
We compare our LBLM model with a representative set of baseline methods, including two fully-supervised CNN-based models (EEGNet~\cite{lawhern2018eegnet} and TCNet~\cite{ingolfsson2020eeg}), two fully-supervised Transformer-based models (EEGConformer~\cite{song2022eeg} and STTransformer~\cite{song2021transformer}), and a pretrained baseline model, LaBraM~\cite{jiang2024large}. For the LaBraM model, we use LaBraM-Base configuration (5.8 M) initialied with the pretrained weight provided from the author. In addition to these external baselines, we perform internal comparisons on two backbone variants. Specifically, we additionally devise LBLM model variants with Transformer backbone (LBLM$_T$) and conformer (LBLM$_C$). For fair comparisons, all models are trained and evaluated using the same EEG data. For the evaluation metrics, we report the accuracy for both word-level and semantic-level classification tasks.

\section{Results}

\begin{table*}[h!]
\renewcommand{\arraystretch}{0.9}
\centering
\caption{Performance on word-level classification task\label{tab:word-level-cross-session}}
\begin{tabular}{l|c|cccccccccccc|c}
\toprule
\textbf{Model} & \textbf{Param.} & \textbf{S01}  & \textbf{S02}  & \textbf{S03}  & \textbf{S04}  & \textbf{S05}  & \textbf{S06}  & \textbf{S07}  & \textbf{S08}  & \textbf{S09}  & \textbf{S10}  & \textbf{S11}  & \textbf{S12}  & \textbf{Avg.} \\ \midrule
EEGNet         & 8.54 K          & 40.8          & 35.3          & 33.5          & \textbf{22.9} & 17.7          & 10.6          & 49.6          & 14.0          & 38.7          & 24.7          & 37.4          & 41.3          & 30.5          \\
TCNet          & 78.62 K         & 36.9          & 36.4          & 32.7          & 22.3          & 14.3          & 13.5          & 44.4          & \textbf{16.4} & 50.4          & 25.5          & 46.8          & 48.1          & 32.3          \\
EEGConformer   & 0.75 M          & 58.2          & 32.7          & 27.5          & 12.2          & 11.2          & 7.0           & 30.9          & 7.0           & 23.9          & 28.3          & 35.1          & 48.3          & 26.9          \\
STTransformer  & 2.78 M          & 60.3          & 16.4          & 30.9          & 21.3          & 13.2          & \textbf{17.4} & 35.1          & 15.6          & 31.2          & 23.4          & 46.2          & 53.8          & 30.4          \\
LaBraM         & 5.80M            & 63.4          & 13.0          & 15.1          & 9.1           & 8.6           & 7.5           & 34.3          & 8.3           & 13.2          & 10.4          & 20.8          & 44.2          & 20.6          \\ \midrule
LBLM           & 22.61 M         & 75.8          & 33.5          & \textbf{36.1} & 22.3          & 16.9          & 11.9          & 47.8          & 10.1          & 47.8          & 17.7          & 47.8          & 55.6          & 35.3          \\
w/ MSTP        & 22.61 M         & \textbf{76.1} & 35.6          & 31.7          & 21.0          & 14.5          & 10.4          & 42.9          & 10.1          & 65.5          & 24.2          & 50.6          & 59.7          & 36.9          \\
w/ MSTP + ASTP & 22.61 M         & 73.8          & \textbf{37.1} & 35.8          & 21.3          & \textbf{18.7} & 11.7          & \textbf{49.9} & 12.7          & \textbf{66.2} & \textbf{32.2} & \textbf{52.7} & \textbf{62.9} & \textbf{39.6} \\
\bottomrule 
\end{tabular}\vspace{-1em}
\end{table*}

\begin{table*}[ht]
\renewcommand{\arraystretch}{0.9}
\centering
\caption{Performance on semantic-level classification task\label{tab:group-level-cross-session}}
\begin{tabular}{l|c|cccccccccccc|c}
\toprule
\textbf{Model} & \textbf{Param.} & \textbf{S01}  & \textbf{S02}  & \textbf{S03}  & \textbf{S04}  & \textbf{S05}  & \textbf{S06}  & \textbf{S07}  & \textbf{S08}  & \textbf{S09}  & \textbf{S10}  & \textbf{S11}  & \textbf{S12}  & \textbf{Avg.} \\ \midrule
EEGNet         & 8.54 K          & 69.4          & \textbf{36.4} & \textbf{38.2} & 31.4          & 23.6          & 33.2          & 46.5          & 30.9          & 18.4          & \textbf{33.5} & 54.3          & 69.4          & 40.4          \\
TCNet          & 78.62 K         & 77.9          & 30.6          & 35.6          & \textbf{31.9} & 22.9          & 30.4          & 56.6          & \textbf{33.5} & 23.4          & 30.3          & 56.9          & 69.4          & 41.6          \\
EEGConformer   & 0.75 M          & 66.2          & 26.2          & 35.3          & 30.4          & 22.3          & 22.9          & 37.9          & 23.9          & 21.6          & 28.8          & 34.5          & 57.9          & 34.0          \\
STTransformer  & 2.78 M          & 65.2          & 25.7          & 27.0          & 28.8          & 20.0          & 22.3          & 36.9          & 26.2          & 39.7          & 27.3          & 57.9          & 74.0          & 37.6          \\
LaBraM         & 5.80 M           & 56.4          & 24.7          & 22.1          & 19.5          & 19.5          & 21.3          & 36.4          & 19.5          & 19.5          & 21.8          & 27.8          & 42.9          & 27.6          \\ \midrule
LBLM           & 22.61 M         & 75.8          & 27.8          & 34.8          & 28.3          & 24.9          & 27.3          & 47.0          & 25.7          & 62.1          & 30.6          & 57.4          & 64.4          & 42.2          \\
w/ MSTP        & 22.61 M         & 76.1          & 27.8          & 36.1          & 28.8          & 26.0          & 29.6          & 47.3          & 26.0          & 72.2          & 29.4          & 57.1          & 81.3          & 44.8          \\
w/ MSTP + ASTP & 22.61 M         & \textbf{79.5} & 28.6          & 36.4          & 29.6          & \textbf{26.5} & \textbf{33.2} & \textbf{59.5} & 27.0          & \textbf{76.4} & 31.2          & \textbf{58.4} & \textbf{77.7} & \textbf{47.0} \\
\bottomrule
\end{tabular}\vspace{-1em}
\end{table*}

\subsection{Classification Performance}
We evaluate the classification performance of the proposed EEG backbone model on two downstream tasks: word-level and semantic-group classification. 

As shown in Table~\ref{tab:word-level-cross-session} and Table~\ref{tab:group-level-cross-session}, our LBLM model outperforms all baselines and achieves the state-of-the-art average accuracy of 39.6\% and 42.8\% on the word-level and semantic-level classification tasks, respectively. Especially in word-level classification, our model achieves a significant improvement of +7.3\% compared to the best baseline TCNet model. These findings align with previous works on pretrained EEG backbone that pre-training yields higher accuracy improvements on fine-grained classification tasks~\cite{jiang2024large,jiang2024neurolm}. In the proposed LBLM model, we observed that both MSTP and FSTP pretraining bring improvements to the model's performance. The MSTP pretraining increased the average accuracy by +1.6\% and +2.6\% on both downstream tasks, while FSTP can further increase the averaged accuracy to 39.6\% (+2.7\%) and 47.0\% (+2.2\%). Notable improvements can be observed on subjects S09 and S10, where our pretraining paradigm increases accuracy from 47.8$\%$ to 66.2$\%$ (S09) and from 17.7$\%$ to 32.2$\%$ (S10) on word-level classification task. These results indicate that the proposed MSTP and ASTP pretraining method is able to capture how brain activity unfolds over time, resulting in more informative features beneficial for downstream classification tasks. 

Compared to baseline convolutional and Transformer models without pretraining, we find that larger Transformer architectures such as EEGConformer and STTransformer produce less generalizable features when trained on a specific participant's data. For both tasks, the participant-specific and averaged accuracies were outperformed by lightweight CNN-based models like EEGNet and TCNet. This aligns with a common observation in EEG model training that smaller models often adapt more easily to task-specific patterns for EEG classification. On the other hand, the pretrained LaBraM performs worst among the baseline. We consider it is mainly because of the lack of time-frequency features learned from their model during the pretraining stage, as well as the fact that it was trained on only 12 EEG channels, which could not handle information from all 122 EEG channels in our dataset. Our LBLM model achieves higher accuracy despite its much larger parameter size because the proposed training paradigm effectively trains the LBLM to capture a diverse range of temporal and spectral features.

\subsection{Ablation Analysis}
\subsubsection{Ablation on Backbone Architecture}
For ablation studies, we first compare LBLM backbone variants: LBLM$_T$ and LBLM$_C$, which use the generic Transformer and Conformer blocks to build the backbone. As shown in Table \ref{tab:loss-ablation-arch}, LBLM$_C$ outperforms LBLM$_T$ on both downstream tasks due to having the convolution module to capture local patterns such as microstates and short-term frequency shifts. Our layer-gating mechanism further improves performance by enabling stable training and flexible feature selection. This ablation justifies our design choice for the LBLM backbone. 
\begin{table}[h!]
\renewcommand{\arraystretch}{0.9}
\centering
\begin{threeparttable}
\caption{Ablation on LBLM backbone variants\label{tab:loss-ablation-arch}}
\begin{tabular}{l|c|c|c}
\midrule
\textbf{Backbone} & \textbf{Param.} & \textbf{Acc.W. (\%)} & \textbf{Acc.S.  (\%)}      \\ \midrule
LBLM$_T$          & 22.36 M          & 25.5                     & 39.8              \\
LBLM$_C$          & 22.37 M         & 31.3 (+5.8)              & 42.7 (+2.9)        \\
LBLM              & 22.61 M         & \textbf{39.6 (+8.3)}     & \textbf{47.0 (+4.3)}   \\ \midrule
\end{tabular}
\begin{tablenotes}
\small
\item Acc.W. and Acc.S. denotes accuracy on word-level and semantic-level classification tasks, respectively.
\end{tablenotes}
\end{threeparttable}\vspace{-1em}
\end{table}
\subsubsection{Ablation on Pretraining Loss}
We further investigate how the loss terms on raw waveform, amplitude, and phase affect LBLM's performance in the pretraining stages and downstream classification tasks. For both MSTP and ASTP pretraining, we evaluate the mean squared error (MSE) and mean absolute error (MAE) of the predicted EEG waves. As shown in Table~\ref{tab:loss-ablation}, adding the amplitude loss ($L^{H}_a$) and the phase loss term ($L^{H}_\phi$) increases MSE and MAE during MSTP and FSTP pretraining. This shows that training the LBLM backbone with multiple prediction heads slightly compromises the prediction performance on EEG waves. However, adding loss terms on the frequency domain ($L^{H}_a$ and $L^{H}_\phi$) also lead to improved performance on both downstream classification tasks. These results suggest that spectro-temporal prediction in FSTP is effective as the addition of spectro prediction terms improves the quality of the EEG representations extracted by the model. 

\begin{table}[ht]   
    \renewcommand{\arraystretch}{0.9} 
\centering
\begin{threeparttable}
\caption{Ablation on pretraining loss terms\label{tab:loss-ablation}}
\begin{tabular}{l|cc|cc|cc}
\toprule
Training                 & \multicolumn{2}{c|}{MSTP} & \multicolumn{2}{c|}{ASTP} & \multicolumn{2}{c}{Classification} \\ \midrule
Mectics                  & MSE         & MAE         & MSE         & MAE         & Acc.W. (\%)      & Acc.S. (\%)    \\ \midrule
$L^{H}_w$                & 0.14        & 0.22        & 0.23        & 0.28        & 37.4             & 42.4\%          \\
+$L^{H}_a$               & 0.15        & 0.24        & 0.34        & 0.30        & 38.5             & 46.1\%          \\
+$L^{H}_a$+$L^{H}_\phi$  & 0.15        & 0.24        & 0.27        & 0.32        & 39.6             & 47.0\%          \\ 
\bottomrule
\end{tabular}
\begin{tablenotes}
\small
\item Acc.W. and Acc.S. denotes accuracy on word-level and semantic-level classification tasks, respectively.
\end{tablenotes}
\end{threeparttable}
\vspace{-1em}
\end{table}

\subsubsection{Ablation on classifier structure}

\begin{table}[h!]
    \renewcommand{\arraystretch}{0.9}
\centering
\begin{threeparttable}
\caption{Ablation result with different classifiers on our pretrained backbone.\label{tab:transfer-learning}}
\begin{tabular}{ll|c|c}
\toprule
\textbf{Classifier}    & \textbf{Training} & \textbf{Acc.W. (\%)} & \textbf{Acc.S. (\%)} \\ \midrule
Linear                 & w/o Pretrain      & 21.2\%               & 30.7                 \\
Linear                 & w/ Pretrain       & 21.9\%               & 31.1                 \\
Convolution            & w/o Pretrain      & 33.8\%               & 39.8                 \\
Convolution            & w/ Pretrain       & 35.5\%               & 42.5                 \\
Spatio-Temporal (Ours) & w/o Pretrain      & 35.3\%               & 42.2\%               \\
Spatio-Temporal (Ours) & w/ Pretrain       & \textbf{39.6\%}      & \textbf{47.0\%}      \\
\bottomrule
\end{tabular}
\begin{tablenotes}
\small
\item Acc.W. and Acc.S. denotes accuracy on word-level and semantic-level classification tasks, respectively.
\end{tablenotes}
\end{threeparttable}\vspace{-1em}
\end{table}
We evaluate the generalizability of our pretrained LBLM backbone on three classifiers. We compare our spatio-temporal (ST) classifier with alternatives used in prior work, including a linear classifier (Pooling + FFN) from EEGConformer/STTransformer and a convolutional classifier (Depthwise + Separable Conv + FFN) from EEGNet. As shown in Table~\ref{tab:transfer-learning}, pretraining consistently improves classification performance across all classifiers. The linear classifier shows only a marginal gain (21.93\% vs. 21.19\%), likely due to the lack of spatial modeling, which limits its ability to leverage the learned representations for silent speech recognition. On the other hand, the convolutional classifier benefits more noticeably, improving from 33.81\% to 35.45\%. These results demonstrate that while different classifiers benefit to varying degrees, the pretrained LBLM backbone consistently enhances downstream performance, with the greatest gains achieved when paired with our ST classifier.

\subsection{Performance of Future EEG Wave Prediction}
Unlike existing EEG pretraining or end-to-end classification methods~\cite{jiang2024large,song2021transformer}, the proposed SFTP pretraining enables our LBLM model to infer future brain states by predicting raw EEG signals, which can provide deeper insights into human cognitive processes \cite{sun2024predicting}.In silent speech active BCI, the capacity to predict speech attempts can help to reduce decoding frequency and hardware cost. On the other hand, our model can help recover missing temporal signals caused by electrode disconnections, enhancing system reliability in real-time applications.
To evaluate future EEG signal prediction, we compare our model against PatchTST~\cite{nie2022time}, a strong baseline for long-term forecasting, and TimesFM~\cite{das2023decoder}, a state-of-the-art foundation model pretrained for zero-shot time-series prediction. We evaluate MSE and MAE on prediction target window size 46 to 136 on unseen EEG segments from our dataset. 
\begin{table}[h!]
    \renewcommand{\arraystretch}{0.9}
\centering
\begin{threeparttable}
\caption{EEG forecasting Performance\label{tab:forecasting}}
\begin{tabular}{l|cc|cc|cc}
\toprule
Model       & \multicolumn{2}{c|}{LBLM   (Ours)} & \multicolumn{2}{c|}{PatchTST} & \multicolumn{2}{c}{TimesFM} \\ \midrule
Mectrics    & MSE              & MAE             & MSE           & MAE           & MSE          & MAE          \\ \midrule
$46(204)$   & 0.27             & 0.32            & 0.84          & 0.62          & 1.40         & 0.76         \\
$70(180)$   & 0.33             & 0.36            & 0.90          & 0.64          & 1.70         & 0.82         \\
$94(156)$   & 0.45             & 0.42            & 0.93          & 0.66          & 2.07         & 0.88         \\
$118(132)$  & 0.50             & 0.44            & 0.98          & 0.68          & 2.49         & 0.91         \\
$136(114) $ & 0.54             & 0.46            & 1.03          & 0.70          & 3.51         & 0.97         \\
\bottomrule 
\end{tabular}
\begin{tablenotes}
\small
\item The number in the brackets denotes the length of the input EEG wave (context window), and the number outside the brackets denotes the length of the predicted EEG wave (target window).
\end{tablenotes}
\end{threeparttable}\vspace{-1em}
\end{table}

\begin{figure}[ht!]
\centering
\includegraphics[width=0.95\linewidth]{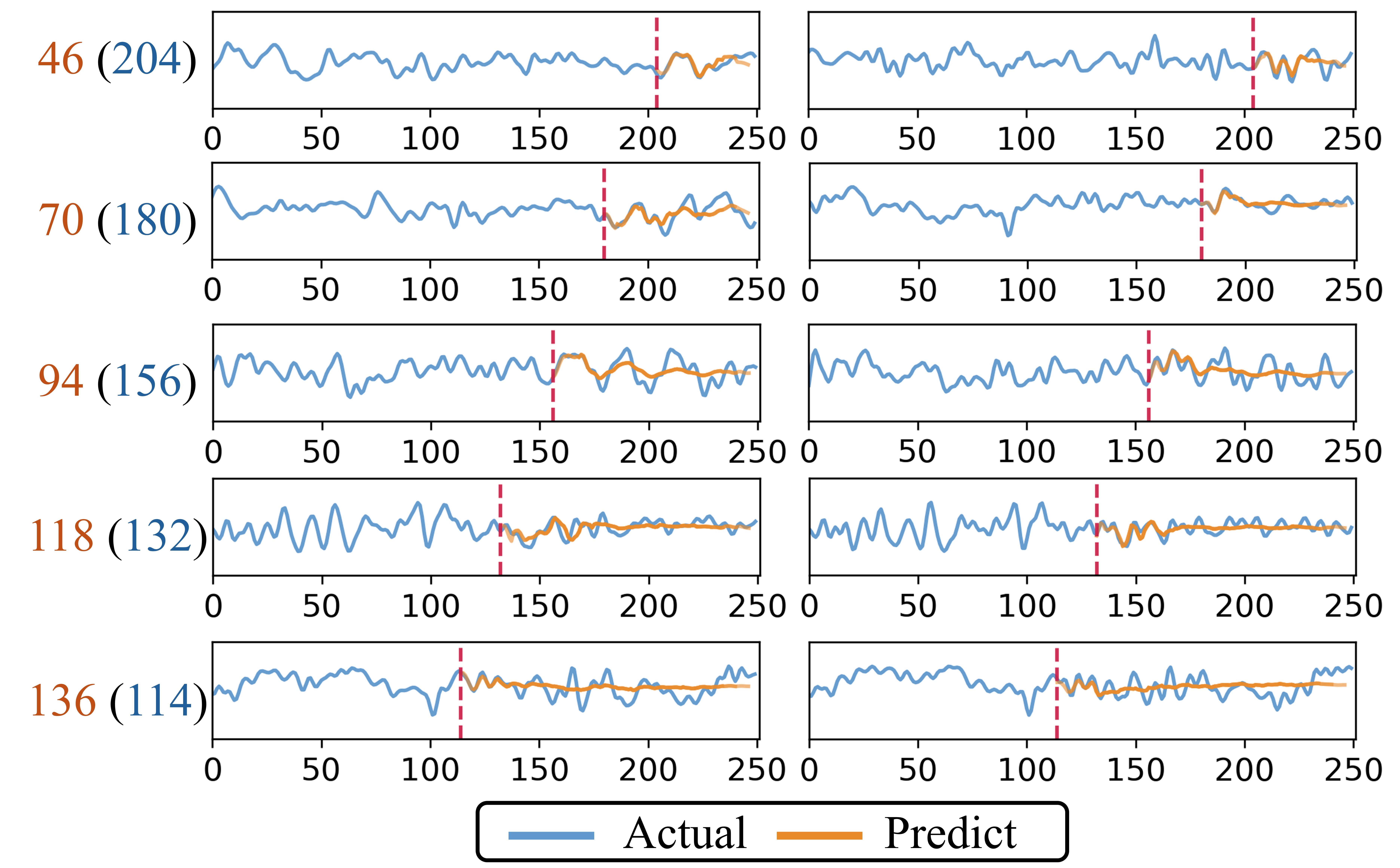}
\caption{Future EEG wave prediction on unseen EEG segments. The blue curves denote the ground truth EEG waves, and the orange curves denote the predicted EEG waves. The red line indicates the start of the prediction window. The number in the brackets denotes context window length while and the number outside the brackets denotes the target window length.  \label{fig:forecast-visualization}
}\vspace{-1em}
\end{figure}

Table~\ref{tab:forecasting} presents the quantitative results of cross-session EEG forecasting. Our LBLM model consistently outperforms PatchTST and the pretrained TimesFM model across all prediction horizons. For instance, with an input EEG length of 204 and a prediction length of 46, LBLM achieves an MSE of 0.27 and an MAE of 0.32, compared to PatchTST's (0.84 and 0.62) and TimesFM (1.40 and 0.76). For our LBLM model, Both MSE and MAE increase as the prediction length increases, suggesting an accumulation of prediction errors that in the autoregressive process. To evaluate performance more visually, we apply post-processing to merge the overlapping EEG predictions from each token and depict the results for target window size 46 to 136 in Figure~\ref{fig:forecast-visualization}. We show 1 second of EEG signals with a 250 Hz sampling rate. We can observe that rougly within the first 30 time steps (120 ms) of the prediction window, the predicted waveforms show strong alignment with the ground truth (blue). However, beyond 120 ms, the LBLM model struggles to follow the complexity, with trends and oscillatory patterns gradually diverging from the true signal. More visualization from the PatchTST and TimesFM model are shown and discussed in Appendix \ref{appendix:visualization}.

\section{Conclusion}
This paper presents a novel self-supervised pretraining approach and introduces the Large Brain Language Model (LBLM), a 22-million-parameter model designed for silent speech decoding in active BCI systems. We collected a new EEG dataset comprising over 120 hours of recordings from 12 participants. To pretrain LBLM, we propose the Future Spectro-Temporal Prediction (FSTP) paradigm, which integrates masked spectro-temporal prediction and autoregressive spectro-temporal prediction to effectively learn from unlabeled EEG data and capture both temporal and spectral representations relevant to downstream classification tasks. We evaluate LBLM against both supervised and pretrained baselines. Experimental results demonstrate that our model consistently outperforms these baselines in word-level and semantic-level classification, particularly in the challenging cross-session setting. Moreover, LBLM is capable of predicting short-term future EEG signals, offering insights into brain dynamics and improving the reliability of real-time active BCI systems. Overall, this work contributes an effective self-supervised learning framework for EEG pretraining, a new EEG dataset for silent speech research, and strong empirical evidence supporting the feasibility of EEG-based silent speech decoding with significant real-world potential.


\clearpage
\bibliographystyle{unsrt}
\bibliography{LBLM_bib}

\clearpage
\appendix
\begin{center}
    \LARGE \textbf{Appendix}
\end{center}

\section{Method Details}
\subsection{Convolution Module in Conformer Blocks}\label{appendix:ConvolutionModule}
\begin{figure}[htbp]
    \centering
    \includegraphics[width=1.0\linewidth]{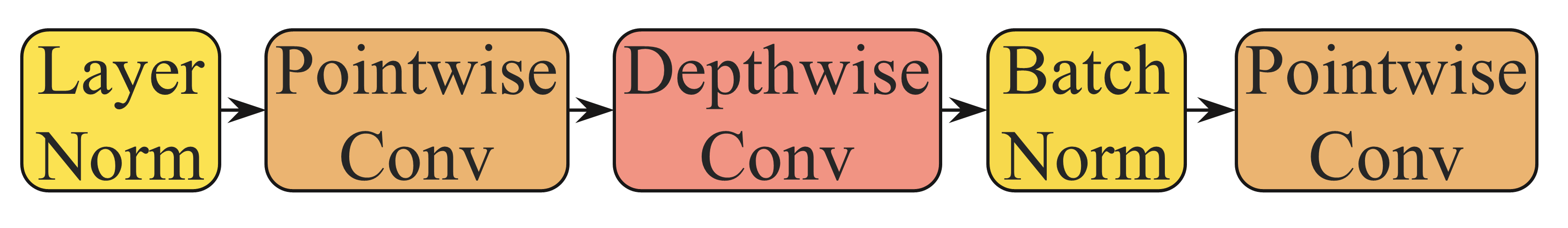}
    \caption{The detailed structure of the convolution module in the LayerGatedConformer blocks.\label{fig:convolution-module}}  
     
\end{figure}

The Conformer blocks we used in our LBLM backbone consist four modules, two feed-forward layer, a convolution module, and a multi-head self-attention (MHSA) layer. The structure of the convolution module is shown in Figure \ref{fig:convolution-module}, which is comprised of two pointwise convolution layers and a depthwise convolution layer. The use of a convolution module inside a Transformer block allows the conformer block to capture local patterns more effectively within each EEG token. 

\subsection{FFT Amplitude and Phase Computation}\label{appendix:fft}
We calculate the FFT amplitude and phase from each EEG patch $\mathrm{p}$ as groundtruth for MSTP and FSTP pretraining. Given a discrete EEG signal $\mathrm{p}$ of length $P$, we use Fast Fourier Transform (FFT) to calculate its frequency components. We first calculate the discrete Fourier transform (DFT) result $X[k]$ of the EEG patch by: 
\begin{equation}
    X[k] = \sum_{n=0}^{N-1} \mathrm{p}[n] \cdot e^{-j \frac{2\pi}{P} kn}, \quad k = 0, 1, \dots, P-1
\end{equation}

Then, based on the freqency powers above, we continue to calculate the amplitude $A[k]$ and the phase $\phi[k]$ by the following equations: 
\begin{equation}
A[k] = |X[k]| = \sqrt{\operatorname{Re}(X[k])^2 + \operatorname{Im}(X[k])^2}
\end{equation}

\begin{equation}
    \phi[k] = \arg(X[k]) = \tan^{-1}\left(\frac{\operatorname{Im}(X[k])}{\operatorname{Re}(X[k])}\right)
\end{equation}

\subsection{Reversible Normalization} \label{appendix:RevIN}
We consider using reversible instance normalization (RevIN) to stabilize the training process and avoid distortions to the original data distribution when outputing the reconstruction and predition of EEG signals. Each EEG channel will be standardized independently across the time axis: 
\begin{equation}
    \hat{\mathbf{x}}^{(i)}_p =  \frac{\mathbf{x}^{(i)}_p - \mu_p^{(i)}}{\sigma_p^{(i)}}
\end{equation}
where $\mu_p^{(i)}$ and $\sigma_p^{(i)}$ denote the mean and variance of a specific EEG channel $i\in \{1,\cdots, M\}$. After model processing, the patch prediction output $\mathbf{y}$ will be de-normalized in a reverse process: 
\begin{equation}
    \hat{\mathbf{y}}_p^{(i)} = \mathbf{y}\cdot\sigma_p^{(i)}+\mu_p^{(i)}
\end{equation}
which ensures the prediction output remains the same distribution as the input. 

\subsection{Multi-band Data Mixing} \label{appendix:Mixing}
\begin{figure}[h!]
\centering
\includegraphics[width=1.0\linewidth]{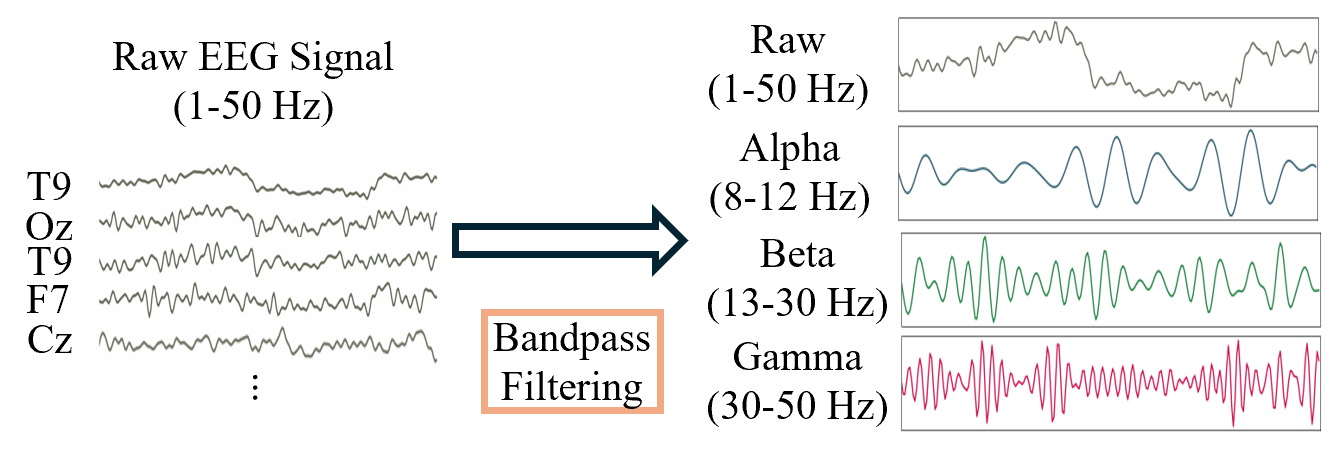}
\caption{Dataset mixing strategy used in this paper. We band-pass filtered the preprocessed raw input EEG data into different frequency bands. Increasing the diversity of the pretraining dataset.\label{fig:mixing_method}}
\end{figure}

Following recent trends in time‐series modeling~\cite{das2024decoder}, where the time-series forecasing model is first trained on simple synthetic data before training on more complex time-series, we also developed a dataset mixing strategy to increase sample size and allow our LBLM model to learn from a more simple waveform at the beginning. However, instead of using sine and cosine waves, we mix the EEG wave components from different band ranges, which is called multi‐band mixing. As shown in Figure \ref{fig:mixing_method}, each raw EEG wave is decomposed into alpha ($8–13\mathrm{Hz}$), beta ($13–30\mathrm{Hz}$), and gamma ($30–50\mathrm{Hz}$) components via bandpass filtering. We select these frequency bands because we consider they are more relevant to silent speech signals and can avoid the model to learn low-frequency drifts or muscle artifacts that is mainly located at frequencies below $4\mathrm{Hz}$. These frequency components are mixed with the raw EEG waves, which we filtered between $1$ to $50\mathrm{Hz}$ as the dataset for pretraining. Empirically, we find the mixing of datasets allows the model update to be more stable, probably due to the more periodic patterns in the time domain makes reconstruct and prediction easier.

\begin{figure}[ht!]
\centering
\includegraphics[width=0.95\linewidth]{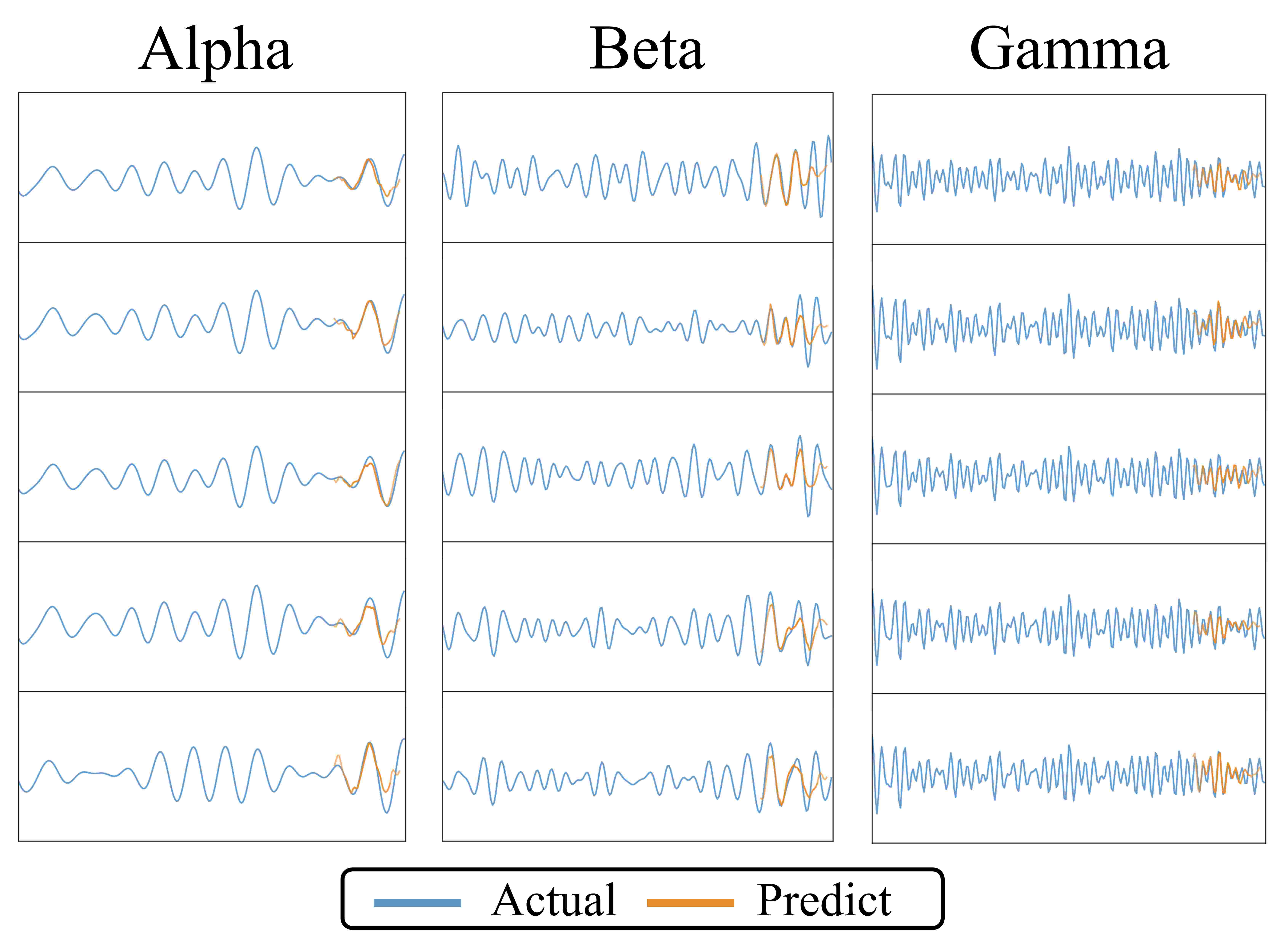}
\caption{EEG signal forecasting visualization of alpha, beta, and gamma band.\label{fig:multiband-visualization}}
\end{figure}
In Figure~\ref{fig:multiband-visualization}, we present future wave prediction results from our LBLM model on the alpha, beta, and gamma bands from several channels. In each sub‐figure, the predicted waveforms (orange) closely align with the actual signals (blue) within a short time period, indicating that our LBLM model can successfully capture both amplitude and finer oscillatory details of the EEG signal from various frequency bands. Notably, based on the past EEG waves, our LBLM model is able to generate both the lower‐frequency fluctuations from the alpha and the higher‐frequency oscillation in beta and gamma bands. Although prediction quality gradually diminishes over longer forecast intervals as the error accumulates, the visualization results indicate the LBLM model's ability to capture the frequency properties solely from raw EEG waves.

\subsection{Finetuning Objective for Classification}\label{appendix:CE}
To finetune the LBLM model for downstream classification tasks, we attach a classifier to the LBLM backbone and train the model with word-level or semantic-level labels using the cross-entropy loss function:  
\begin{equation}
\mathcal{L}(\hat{\mathbf{y}}, \mathbf{y})
  = - \sum y_{c} \, \log\bigl(\hat{y}_{c}\bigr),
\label{eq:crossentropy}
\end{equation}
where $y_{c}$ and $\hat{y}_{c}$ denote the ground truth and the predicted word/semantic label from the EEG segment. 

\subsection{Hyperparameters}\label{appendix:hyperparameter}
Our experiments are implemented using the PyTorch framework and executed on NVIDIA H100 GPUs. During backbone pretraining, we use a batch size of $1024$ and an initial learning rate of $1e^{-3}$. Optimization is performed using the LAMB optimizer~\cite{you2019large}, along with a cosine annealing learning rate scheduler. We set the weight decay to 0.01 and apply gradient clipping with a maximum norm of 1.0 to ensure training stability. For the Huber loss, we use $delta=1$ to balance between the MSE and MAE terms. Additionally, we set $\lambda_1=\lambda_2=0.1$ to balance the learning of amplitude and phase components in the EEG signals. The LBLM backbone model is pretrained for $80$ epochs during MSTP and FSTP task before finetuning to a subject specific dataset. Our LBLM model is initilized with 4 conformer layers with the layer-gating. The hidden dimension and the feedforward dimensions are set to 64.

\section{EEG data collection}\label{sec:dataset-collection}
\begin{figure}[ht!]
    \centering
    \includegraphics[width=0.95\linewidth]{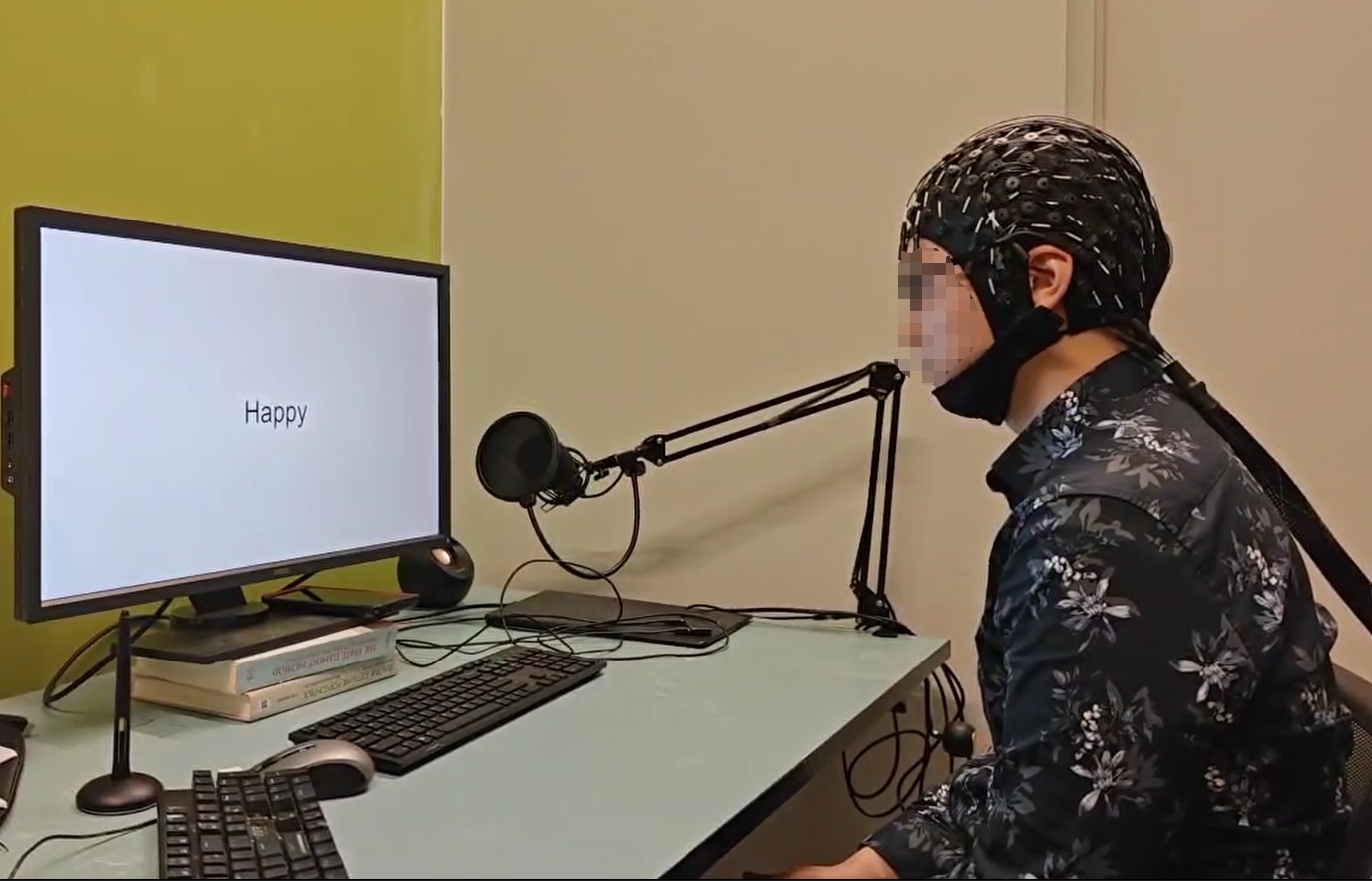}
    \caption{Silent speech experiment environment. Participants were comfortably seated on a chair in front of a monitor.  \label{fig:photo}}
    \end{figure}

Twelve right-handed native English speakers from Australia ($6$ males, $6$ females, aged $28 \pm 0.85$ years old) with no visual or speech impairment or neurological disorders were participated in this study. They performed silent speech task during the experiment. The subjects were seated on a comfortable chair and were instructed to keep their eyes open while observing a computer monitor placed at a comfortable distance. Subjects were also asked to keep their body movement o minium during the experiment to minimise muscle. Informed consent was obtained in writing from each participant prior to the experiment. The EEG signals were recorded at a sampling rate of $1000\mathrm{Hz}$ using a 128-channel neuroscan quik-cap\footnote[1]{https://au.neuromedicalsupplies.com/product/compumedics-neuroscan-quik-cap-128-ch-large-62-68cm-synamps2-agagcl-sintered-electrodes-2/} connected to a SynAmps RT 128-channel Amplifier\footnote[2]{https://compumedicsneuroscan.com/product/synamps-rt-128-channel-eeg-erp-ep-amplifier/}. The CURRY 9 software\footnote[3]{https://compumedicsneuroscan.com/product/curry-9-x-data-acquisition-and-online-processing/} was used for data acuisition. The placement of electrodes follows the 10–10 international system~\cite{jasper1958ten}. Impedance checks were conducted before each experiment to ensure that the impedance value of all electrodes remains below $10k\omega$ and balanced across electrodes~\cite{gorecka2019dependence, valle2024identification}. A ground electrode was referenced to maintain signal consistency. The setup of the EEG cap as well as the monitor in the lab is shown in Figure \ref{fig:photo}.

\subsection{Experiment Design}
Our silent speech experiment involves $24$ commonly used words in English. The words are selected based on 6 semantic groups. Table \ref{Tab:ds-stim} shows the 24 words and their corresponding semantic group. Since our goal is to facilitate silent speech decoding for active BCI research, we include a total of $6$ semantic groups ranging from motion, emotion, location, person, object, and number to increase vocabulary diversity. The participants were instructed to make articulation attempts without emitting sounds.  Each word will be repeated $5$ times and the sequence of the words is randomized for each session. The experiment was implemented in Python using Psychopy library. Statistical information of the collected dataset is shown in Table~\ref{Tab:ds-summary}. 

\begin{table}[h]
    \centering
    \caption{Summary of Data Collection Settings\label{Tab:ds-summary}}
    \begin{tabular}{c|c}
    \toprule
    Feature         & Details \\\midrule
    Subjects        & 12      \\
    Modality        & EEG     \\
    Channel Number  & 122     \\
    Sessions Per Subject        & 16      \\
    Total Trials Per Subject   & 6000    \\
    Vocabulary Size & 24      \\
    Total Hour Per Subject     & 10     \\ 
    \bottomrule 
    \end{tabular}
\end{table}

\begin{table*}[h!]
    \centering
    \caption{List of stimuli and their syllable number used in our experiment\label{Tab:ds-stim}}
    \begin{tabular}{ccc|ccc}
    \toprule
    Semantic Category         & Word        & Number of Syllables & Semantic Group           & Word      & Number of Syllables \\
    \multirow{4}{*}{Motion}    & Jumping     & 2         & \multirow{4}{*}{People}  & Mother    & 2         \\
                              & Running     & 2         &                          & Cowboy    & 2         \\
                              & Swimming    & 2         &                          & Professor & 3         \\
                              & Going       & 2         &                          & Me        & 1         \\
    \multirow{4}{*}{Emotion}  & Happy       & 2         & \multirow{4}{*}{Number}  & One       & 1         \\
                              & Sad         & 1         &                          & Three     & 1         \\
                              & Fun         & 1         &                          & Eleven    & 3         \\
                              & Horrible    & 3         &                          & Million   & 2         \\
    \multirow{4}{*}{Location} & College     & 3         & \multirow{4}{*}{Object} & Spoon     & 1         \\
                              & Home        & 2         &                          & Alfa      & 2         \\
                              & Battlefield & 3         &                          & Python    & 2         \\
                              & Here        & 2         &                          & Telephone & 3         \\ 
    \bottomrule
\end{tabular}
\end{table*}

\subsection{EEG data preprocessing}\label{sec:preproc}
\begin{figure}[h!]
    \centering
    \includegraphics[width=0.95\linewidth]{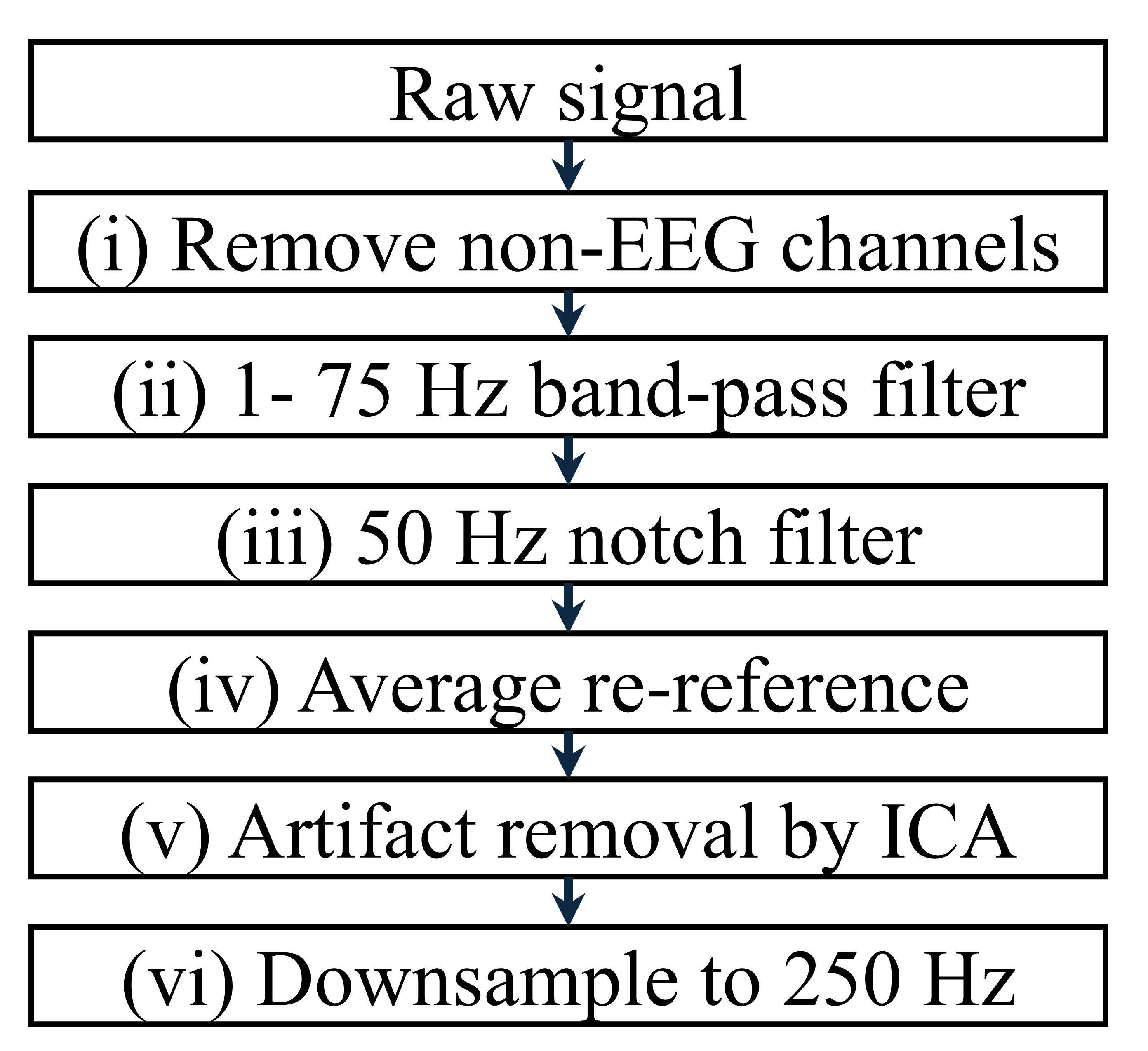}
    \caption{Pre-processing steps for EEG data. \label{fig:preprocessing} }
\end{figure}

An overview of the preprocessing pipeline is shown in Figure \ref{fig:preprocessing}. Using our 128-channel neuroscan system, each EEG recording is saved in $.cdt$ format and consists of 133 channels: 122 EEG channels capturing brain activity, 4 electrodes capturing eye movements, and 6 reference clectrodes capturing muscle and environment noise, and 1 trigger channel for condition labeling. The EEG data is first band-pass filtered between $1$ and $75\mathrm{Hz}$ using a finite impulse response (FIR) filter to remove slow drifts and high‐frequency noise, followed by a $50\mathrm{Hz}$ notch filter to eliminate line interference. Then the EEG signals are re-referenced by calculating the average of the 122 channels. Next, ICA decomposition is performed to identify and remove artifact components such as eye movement, muscle activity, and cardiac artifacts, with a confidence threshold of 90\% for ICA classification. Finally, the EEG data is segmented into epochs of 2 seconds with a 0.5-second overlap. When used for training and classification, the EEG data is downsampled to $250\mathrm{Hz}$ to improve computational efficiency. Our preprocessing pipeline is implemented using the EEGLAB library~\cite{delorme2004eeglab}.  

\begin{figure}[h!]
    \centering
    \includegraphics[width=0.95\linewidth]{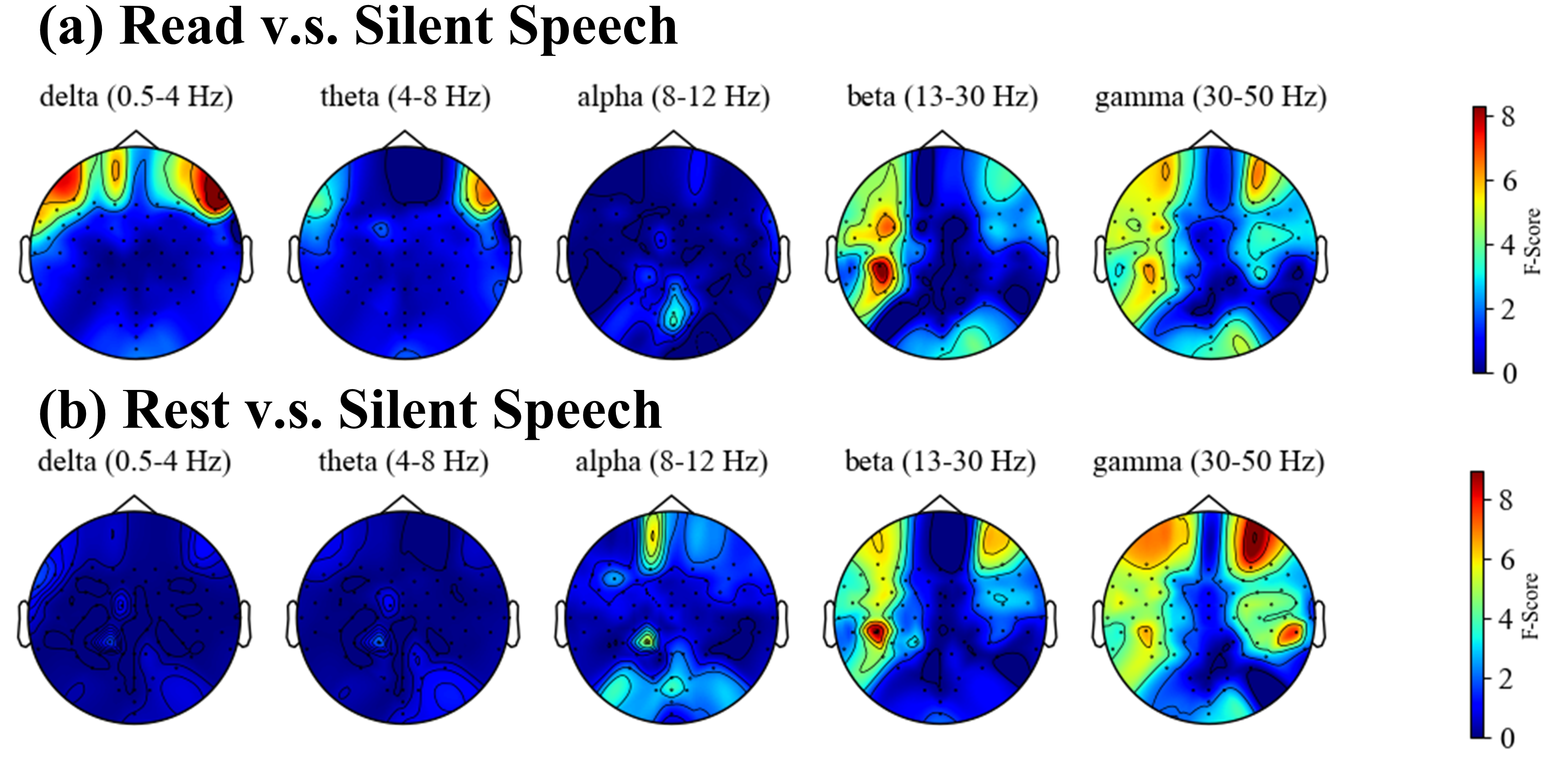}
    \caption{
    rm-ANOVA analysis of paired conditions from all subject. (a) read vs. silent speech, (b) rest vs. silent speech. We visualize the F-score of the compared conditions. \label{fig:rmanova} }
\end{figure}

\begin{figure*}[ht!]
    \centering
\includegraphics[width=0.95\linewidth]{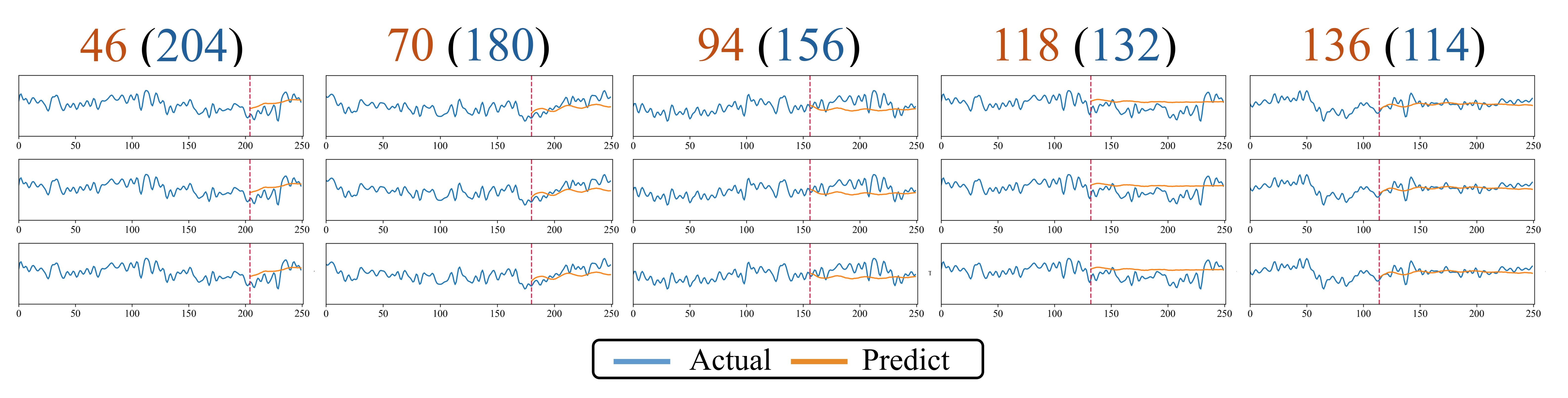}
\caption{
    Future EEG wave prediction on unseen EEG segments from PatchTST model. The blue curves denote the ground truth EEG waves, and the orange curves denote the predicted EEG waves. The red line indicates the start of the prediction window. The number in the brackets denotes context window length while and the number outside the brackets denotes the target window length. \label{fig:PatchTST_FuturePrediction}}
\end{figure*}

\begin{figure*}[ht!]
\centering
\includegraphics[width=0.95\linewidth]{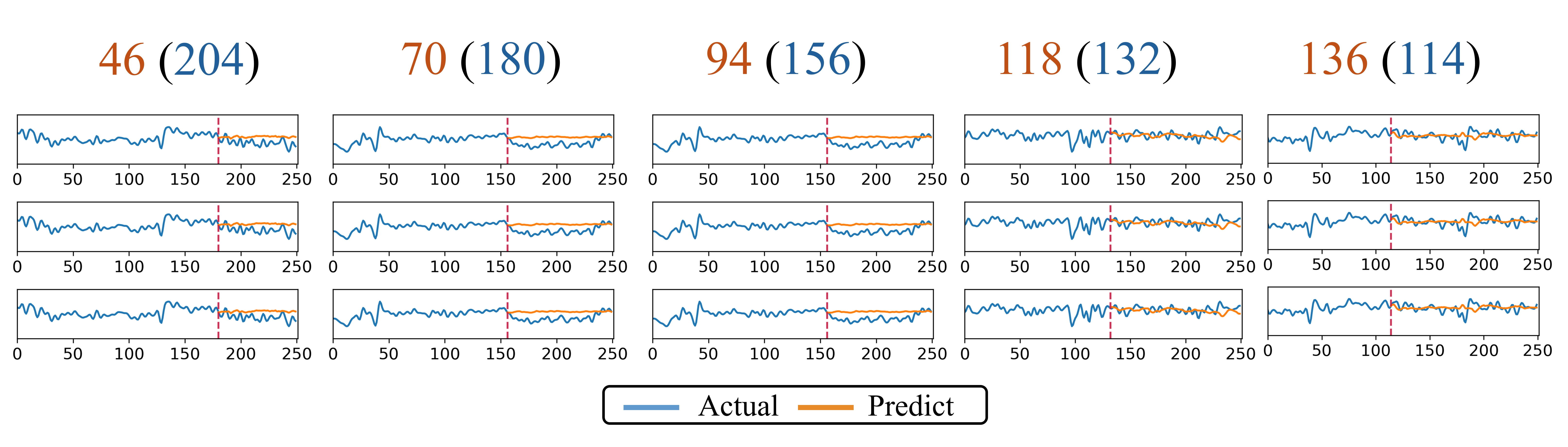}
\caption{
    Future EEG wave prediction on unseen EEG segments from TimesFM model. The blue curves denote the ground truth EEG waves, and the orange curves denote the predicted EEG waves. The red line indicates the start of the prediction window. The number in the brackets denotes context window length while and the number outside the brackets denotes the target window length. \label{fig:TimesFM_FuturePrediction}}
\end{figure*}

\begin{figure}[ht!]
    \centering
    \includegraphics[width=0.95\linewidth]{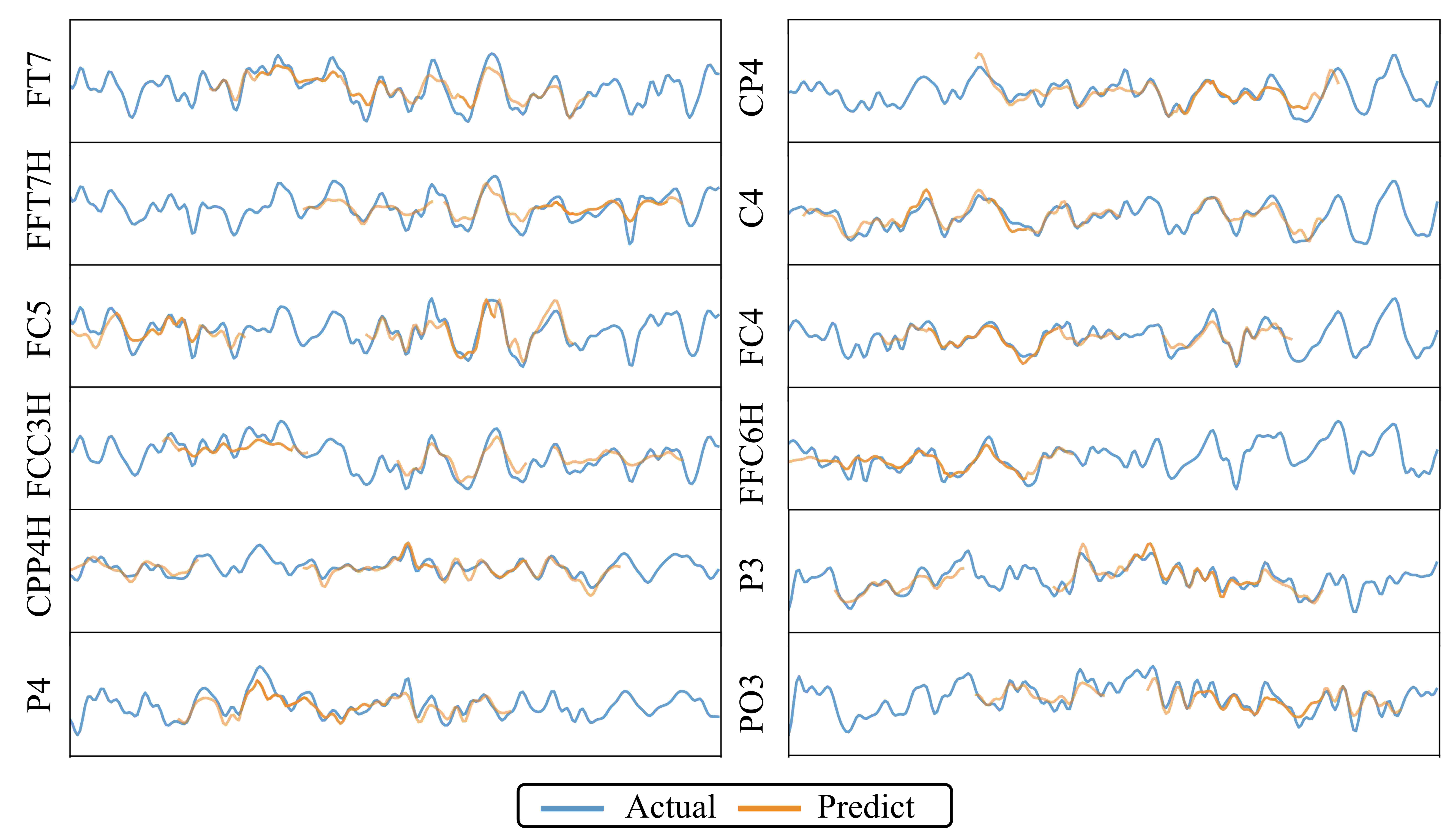}
    \caption{
    EEG signal masked reconstruction visualization of 12 channels. The blue curves are the actual data. The curves before the black dashed lines are the actual input to the model, and the orange curves are the predicted EEG signal.\label{fig:masked-visualization}}
\end{figure}

\subsection{Correlation Analysis}\label{sec:anova}
To determine which frequency bands exhibited the greatest variation under different cognitive conditions, we conducted a repeated‐measures analysis of variance (rm‐ANOVA) on the power spectrum density (PSD) of delta ($1-4\mathrm{Hz}$), theta ($4-8\mathrm{Hz}$), alpha ($8-13\mathrm{Hz}$), beta ($13-30\mathrm{Hz}$), and gamma ($30-50\mathrm{Hz}$). Three condition pairs were compared: rest versus reading, rest versus silent speech, and reading versus silent speech. We used F‐scores to quantify differences among these conditions, with higher F‐scores indicating larger variations in PSD. 
Figure~\ref{fig:rmanova}(a) shows that in the delta band (1–4 Hz), strong frontal activation likely corresponds to eye‐related artifacts, while the alpha band (8–13 Hz) reveals moderate differences centered around the occipital region, notably near the POOZ electrode. In contrast, the beta (13–30 Hz) and gamma (30–50 Hz) bands exhibit high F‐scores in the left hemisphere, suggesting substantial neural variability that may be related to motor or cognitive processes. On the other hand, Figure~\ref{fig:rmanova}(b) highlights the beta and gamma bands in the left hemisphere as the main source of variation when comparing resting versus silent speech. The relative lack of occipital activation here contrasts with the reading condition, indicating reduced visual processing during silent speech and a stronger emphasis on cognitive or motor‐planning regions. Based on these findings, alpha, beta, and gamma bands emerge as the most discriminative for silent speech decoding, justifying our focus on these frequency ranges in subsequent analyses. Altogether, these analysis results reveal that alpha, beta, and gamma bands contains particularly significant differences during silent speech activities. This observation motivated our selection of these bands for multi‐band data mixing in subsequent analyses.

\section{Visualizations}\label{appendix:visualization}
\subsection{Reconstruct Performance during MSTP pretraining}
We visualize the masked reconstruction reuslts of LBLM during MSTP pretraining in Figures \ref{fig:masked-visualization}. Here we randomly masked out 10\% of the input tokens. We can see that for most reconstructed patches, the LBLM model is able to preserve the signal morphology and voltage values. When comparing the visualization of the MSTP pretraining to that of the ASTP pretraining, we can see that although the masked reconstruction results are good, the visualization for future EEG timeseries prediction remains low. This result further demonstrate the drawback of current masked-reconstruction-based method where the EEG encoder may not be learning meaningful feature that help understand the EEG signal dynamics.

\subsection{Future EEG Prediction Performance from PatchTST and TimesFM Models}
We further show visualization of future EEG prediction from the PatchTST model pretrained on our dataset and the TimesFM model that is pretrained on large time-series dataset in Figure \ref{fig:PatchTST_FuturePrediction} and Figure \ref{fig:TimesFM_FuturePrediction} respectively. We can see that although the TimesFM model is trained on massive amount of time-series datasets, patterns learned from other time-series data could not generalize to predict the non-stationary EEG signals. On the other hand, pretraining the PatchTST model on our data helps the model to capture trends to a very limited extend, indicating that lacking more local feature extraction methods, a Transformer backbone still struggles to learn and capture the temporal dynamics from EEG data.

\end{document}